\documentclass[10pt,twocolumn,letterpaper]{article}

\usepackage[pagenumbers]{iccv} 

\definecolor{iccvblue}{rgb}{0.21,0.49,0.74}
\usepackage[pagebackref,breaklinks,colorlinks,allcolors=iccvblue]{hyperref}
\usepackage{multicol}
\usepackage{multirow}
\usepackage{pifont}         
\usepackage{amssymb}
\usepackage[frozencache,cachedir=.]{minted}
\usepackage[export]{adjustbox}
\usepackage{fancyvrb}
\usepackage{xcolor}
\usepackage{array}
\usepackage{multicol}
\usepackage{multirow}
\usepackage{tcolorbox}
\usepackage{lipsum} 
\usepackage{fbox}
\usepackage{algpseudocode}
\usepackage{algorithm}

\newcommand{\cmark}{\ding{51}}
\newcommand{\xmark}{\ding{55}}

\definecolor{cvprblue}{rgb}{0.21,0.49,0.74}
\definecolor{turquoise}{rgb}{1, 1, 0.53}
\definecolor{limegreen}{rgb}{1, 0.77, 0}
\definecolor{goldenrod}{rgb}{0.25, 0.88, 0.82}
\definecolor{coral}{rgb}{0.12, 0.59, 0.99}
\newminted{python}{%
    frame=single,
    fontsize=\scriptsize,
    baselinestretch=0.8,
    breaklines,
    breakanywhere,
    style=friendly,
    escapeinside=@@
}
\newcommand{\highlight}[2]{\setlength{\fboxsep}{0pt}\colorbox{#1}{\strut#2}}

\def\paperID{12615} 
\def\confName{ICCV}
\def\confYear{2025}

\title{\modelname: Reverse Engineering CAD Code from Point Clouds}

\author{
    Danila Rukhovich\textsuperscript{1} \\
    {\tt\small danila.rukhovich@uni.lu} \and
    Elona Dupont\textsuperscript{1} \\
    {\tt\small elona.dupont@uni.lu} \and
    Dimitrios Mallis\textsuperscript{1} \\
    {\tt\small dimitrios.mallis@uni.lu} \and
    Kseniya Cherenkova\textsuperscript{12} \\
    {\tt\small kseniya.cherenkova@uni.lu} \and
    Anis Kacem\textsuperscript{1} \\
    {\tt\small anis.kacem@uni.lu} \and
    Djamila Aouada\textsuperscript{1} \\
    {\tt\small djamila.aouada@uni.lu} \and
    {\textsuperscript{1}SnT, University of Luxembourg \quad \textsuperscript{2}Artec3D, Luxembourg}
}

\begin{document}
\newcommand{\modelname}{\texttt{CAD-Recode}}
\maketitle

\begin{abstract}

Computer-Aided Design (CAD) models are typically constructed by sequentially drawing parametric sketches and applying CAD operations to obtain a 3D model. The problem of 3D CAD reverse engineering consists of reconstructing the sketch and CAD operation sequences from 3D representations such as point clouds. In this paper, we address this challenge through novel contributions across three levels: CAD sequence representation, network design, and training dataset. In particular, we represent CAD sketch-extrude sequences as Python code. The proposed \textbf{\modelname}~translates a point cloud into Python code that, when executed, reconstructs the CAD model. Taking advantage of the  exposure of pre-trained Large Language Models (LLMs) to Python code, we leverage a relatively small LLM as a decoder for \modelname~and combine it with a lightweight point cloud projector. \modelname~is trained on a procedurally generated dataset of one million CAD sequences. \modelname~significantly outperforms existing methods across the DeepCAD, Fusion360 and real-world CC3D datasets. Furthermore, we show that our CAD Python code output is interpretable by off-the-shelf LLMs, enabling CAD editing and CAD-specific question answering from point clouds.

\end{abstract}

\section{Introduction}

\begin{figure}[ht]
\begin{center}
\includegraphics[width=\linewidth]{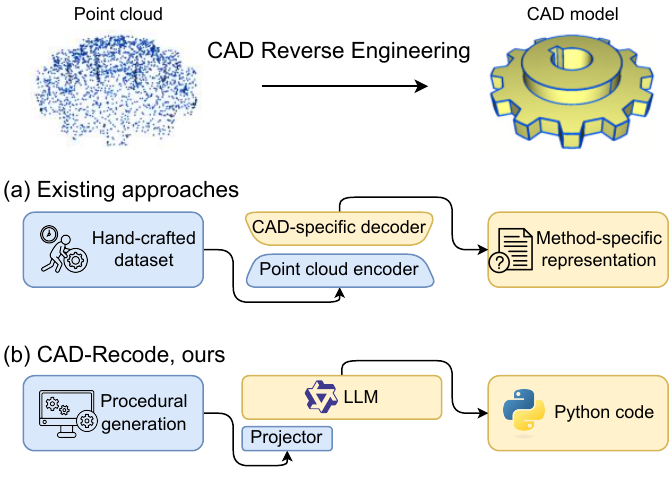}
\end{center}
 \caption{3D CAD reverse engineering is a process of converting a point cloud into a CAD model (top). Existing methods are constrained by the use of method-specific CAD representations and limited hand-crafted training datasets (a). On the contrary, \modelname~employs a pre-trained LLM with a lightweight projector that translates point clouds into executable Python code and is trained on a procedurally generated dataset (b).}
\label{fig:teaser}
\end{figure}

Computer-Aided Design (CAD) modeling is the standard approach for designing manufactured objects, ranging from furniture to mechanical components~\cite{briere2012comparing,liu2020fast}. However, creating a 3D CAD model is a time-consuming task that demands specialized expertise, as the model must not only capture the object's shape but also maintain its functional requirements—commonly referred to as the \textit{design intent}~\cite{li2010detecting, camba2016parametric}. To streamline this process, \textit{3D CAD reverse engineering} aims at generating CAD models directly from 3D scanned objects, offering a faster and more accessible pathway to CAD creation~\cite{helle2021case}. 

Automated 3D CAD reverse engineering has a long history in the fields of computer vision and graphics~\cite{sobh1994industrial, eggert1998simultaneous}, with goals evolving alongside advancements in the field. These objectives have progressed from identifying CAD parts in 3D point clouds~\cite{sobh1994industrial} to 
predicting the sequence of steps a designer may take to recreate a 3D scanned object in CAD software~\cite{ma2024draw,khan2024cad}. This latter goal is particularly appealing, as it aims not only to produce a final CAD parametric model but also to capture the design steps behind it, enabling further editing within CAD software~\cite{uy2022point2cyl,khan2024cad}. In CAD software, designers typically construct their CAD model as feature-based design sequences, where a sequence of 2D sketches is transformed into 3D objects via operations such as extrusion and revolution~\cite{xu2021zonegraph,willis2021fusion}. Following the release of large CAD datasets~\cite{koch2019abc,willis2021fusion,cc3d}, recent works have focused on learning feature-based CAD sequences from input point clouds, specifically as \textit{sketch-extrude} sequences~\cite{wu2021deepcad,uy2022point2cyl,ren2022extrudenet,li2023secad,multicad,khan2024cad,dupont2024transcad}. 
As depicted in~\Cref{fig:teaser}(a), although varying in methodology, these approaches share a common pipeline: (1) crafting a CAD sketch-extrude sequence representation, (2) converting raw CAD data~\cite{koch2019abc,willis2021fusion} into this format, and (3) training dedicated neural networks to output these representations from input point clouds. 

Despite recent advances in feature-based CAD reverse engineering, key limitations constrain the broader applicability of existing approaches. 
Firstly, existing methods often use customized CAD representations, such as custom CAD languages~\cite{wu2021deepcad,khan2024cad,dupont2024transcad,ma2024draw} or simplified extrusion shapes~\cite{ren2022extrudenet,li2023secad,uy2022point2cyl}, to facilitate model training. These representations are difficult to interpret, require post-processing to be compatible with CAD tools, and restrict design capabilities to basic operations. 
Secondly, these approaches typically rely on designing networks that output language-like CAD representations~\cite{khan2024cad,dupont2024transcad} and training them from scratch. This requires the networks to learn not only the geometry of the point clouds, but also the syntax of the CAD sequence representation.

In this paper, we pose the following question: \textit{In view of the recent breakthrough performance of Large Language Models (LLMs), how can their advanced language understanding be leveraged for CAD reverse engineering?} 

To address this question, we base our approach on three key observations: (1) LLMs can generate valid Python code~\cite{yang2024qwen2technicalreport,openai2024gpt4technicalreport}, (2) modern CAD software increasingly supports modeling through Python scripting~\cite{cadquery}, and (3) recent efforts have shown that LLMs can be fine-tuned to process point clouds~\cite{xu2024pointllm,zhang2022pointclip}. As shown in~\Cref{fig:teaser}(b), \textit{we propose \modelname, a solution for CAD reverse engineering by fine-tuning an LLM to map input point clouds into CAD sketch-extrude sequences represented as Python code.} In particular, instead of crafting a CAD representation, we base our representation on the existing Python CadQuery library~\cite{cadquery}. This code-based representation is not only interpretable but also inherently allows for incorporating modular CAD features and design practices such as reusing design elements and abstracting low-level design steps (\eg 3D box to represent a four-line sketch of a square and its extrusion). To learn the mapping between point clouds and CAD Python code, we fine-tune a pre-trained LLM, Qwen2-1.5B~\cite{yang2024qwen2technicalreport}, augmented with a lightweight, trainable point cloud projector. To train \modelname, a potential approach could be using existing sketch-extrude datasets~\cite{wu2021deepcad,willis2021fusion} and converting them to Python CadQuery code. However, these datasets are limited in size and design features included due to the efforts required to convert their original proprietary representation into one that is suitable for learning. As a solution, we propose a \textit{procedurally} generated training dataset composed of one million CAD sketch-extrude sequences as Python CadQuery code. This dataset consists of CadQuery Python scripts generated following predefined heuristics with randomized parameter selection. The execution of each generated script results in a parametric CAD model. Unlike existing CAD datasets, this procedurally generated dataset provides an alternative for learning the mapping between point clouds and CAD sketch-extrude sequences in Python code, with full control over the design features, patterns and dataset size included during training. Our contributions can be summarized to:

\begin{itemize}
\item A CAD sketch-extrude sequence representation in Python code using CadQuery~\cite{cadquery} for CAD reverse engineering. 

\item \modelname, the first LLM-based CAD reverse engineering model designed to predict CAD Python code from point clouds. The model, consisting of a pre-trained LLM and a point cloud projector is trained end-to-end to generate code that reconstructs the input geometry.

\item A one million procedurally generated training dataset of CAD sketch-extrude sequences as CadQuery Python code. This provides precise control over dataset size, features, and design patterns included during training, resulting in significant performance improvement. We will make this dataset publicly accessible.

\item Extensive experiments on three publicly available datasets show that \modelname~achieves substantial improvements over state-of-the-art methods in CAD reverse engineering. Moreover, we show that \modelname, when operating on point clouds and generating CAD code, can be integrated with an off-the-shelf LLM to perform CAD Question Answering (CAD-QA) and CAD editing from point clouds.
\end{itemize}

\noindent \textbf{Paper Organization:} The rest of the paper is organized as follows. Section \ref{sec:related_works} reviews related works. Section~\ref{sec:cad_code} introduces the CAD code representation and the proposed synthetic dataset. \modelname~is formulated and described in Section~\ref{sec:cad recode}.  Experiments are presented in Section~\ref{sec:experiments}. Finally, conclusions and future works are given in Section~\ref{sec:conclusion}.

\section{Related Works} \label{sec:related_works}

\vspace{0.2cm}
\noindent \textbf{LLM, Point Cloud and CAD:}
Recent works have explored integrating point clouds with LLMs for various tasks, including 3D generation~\cite{zhao2023michelangelo, yin2023shapegpt}, captioning~\cite{han2024onellm, xu2024pointllm, 3dllm}, and question answering~\cite{hong20233d,chen2024ll3da}. These approaches typically employ complex point cloud encoders, either aligning with CLIP embeddings~\cite{zhang2022pointclip, Zhu_2023_ICCV, xue2023ulip, liu2024openshape, Xue_2024_CVPR} or directly with LLM feature spaces~\cite{xu2024pointllm}. Such methods require two-stage training: first pre-training the point cloud encoder, then fine-tuning with the LLM through instruction-based prompts. 

In parallel, recent works have started exploring LLMs' capabilities in a range of CAD-related tasks. 
ReparamCAD~\cite{Kodnongbua2023reparamCAD} infers shape variations from parametric models and text descriptions, while CADTalk~\cite{yuan2024cadtalk} generates semantic descriptions of CAD parts. 
The works in~\cite{badagabettu2024query2cad, alrashedy2025generating} focus on the generation of CAD models from text using LLMs, and SGP-Benchmark~\cite{qiu2024can} evaluates LLMs' understanding of CAD sketch-extrude sequences using CAD-specific question answering. While Img2CAD~\cite{you2024img2cad} attempts CAD reverse engineering from images using GPT-4V~\cite{openai2024gpt4technicalreport}, it still requires a separate transformer for parameter inference. 
In contrast, \modelname~introduces the first approach for point cloud to CAD reconstruction combining point clouds with the sequence modeling capabilities of pre-trained LLMs.

\vspace{0.2cm}
\noindent \textbf{CAD Reverse Engineering from Point Cloud:}
CAD reverse engineering aims to reconstruct parametric CAD models from 3D shapes (\eg, meshes or point clouds) in a compatible representation with CAD software. A key challenge lies in the choice of this representation. A line of works attempts to address sub-problems of the CAD reverse engineering task by focusing on parameter estimation for edges and surface primitives~\cite{liu2021pc2wf, cherenkova2023sepicnet, wang2020pienet, sharma2020parsenet, zhu2023nerve, li2019spfn, guo2022complexgen, Cherenkova2024SpelsNet} or reconstructing B-Rep construction history~\cite{willis2021fusion, lambourne2021brepnet, dupont2022cadops, xu2021zonegraph}. In order to obtain a representation that is closer to CAD modelling, several methods~\cite{li2024sfmcad, yu2023d, friedrich2019optimizing, kania2020ucsg, yu2024d} use Constructive Solid Geometry (CSG), representing models as compositions of 3D primitives and Boolean operations. While this enables reconstruction, the CSG representation diverges from modern CAD workflows~\cite{xu2021zonegraph}.

Recent works have adopted the more CAD-aligned sketch-extrude representation, introduced by DeepCAD~\cite{wu2021deepcad} for CAD generation~\cite{xu2023hnc, xu2022skexgen} or predicting extrusion cylinder~\cite{uy2022point2cyl, ren2022extrudenet}. Considering the sequential nature of sketch-extrude operations, methods have explored a template-based approach~\cite{lambourne2022reconstructing} given a rounded voxel input representation. Furthermore, transformer architectures have been investigated for both autoregressive~\cite{khan2024cad} and non-autoregressive~\cite{wu2021deepcad,dupont2024transcad} prediction of sketch-extrude sequences from point clouds. The work in~\cite{ma2024draw} combines a lightweight pre-trained language model~\cite{sanh2019distilbert} with a point cloud encoder using a diffusion-based approach. Alternative methods using self-supervised~\cite{li2023secad} or unsupervised~\cite{li2024sfmcad} learning still face integration challenges due to their non-standard sketch representations (\eg, signed distance functions). In contrast to these approaches that require full parameter learning of specialized networks for both CAD geometry and representation syntax, we leverage pre-trained LLMs that have been exposed to programming patterns through large-scale training on code repositories. Our method outputs Python code using the CadQuery library~\cite{cadquery} that is directly executable and can easily be interpreted. Additionally, we address the data limitation through automated generation of a large-scale training dataset, enabling full control over design features included during training.

\section{CAD Representation as Code} \label{sec:cad_code}

\begin{figure}[ht]
    \setlength{\tabcolsep}{0pt}
    \small
    \begin{tabular}{ccccccc}
    \includegraphics[width=0.221\linewidth]{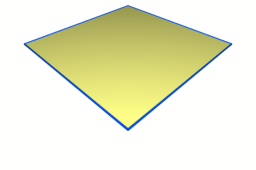} & &
    \includegraphics[width=0.221\linewidth]{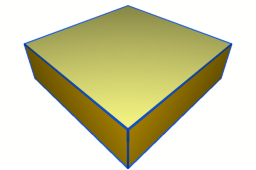} & &
    \includegraphics[width=0.221\linewidth]{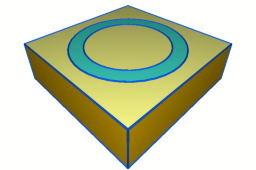} & &
    \includegraphics[width=0.221\linewidth]{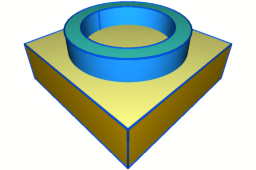} \\
    Sketch 1 & \textrightarrow & Extrude 1 & \textrightarrow & Sketch 2 & \textrightarrow & Extrude 2 \\
    \end{tabular}
    
    \definecolor{turquoise}{rgb}{1, 1, 0.53}
    \definecolor{limegreen}{rgb}{1, 0.77, 0}
    \definecolor{goldenrod}{rgb}{0.25, 0.88, 0.82}
    \definecolor{coral}{rgb}{0.12, 0.59, 0.99}
    \fvset{listparameters=\setlength{\topsep}{0pt}\setlength{\parsep}{0pt}}
     \vspace{-0.5em}
        \begin{center} \small (a) Sketch-Extrude sequence \end{center}
    \vspace{-0.5em}
    \begin{Verbatim}[commandchars=\\\{\}, fontsize=\footnotesize]
\highlight{turquoise!50}{SOL Line -50 -50 Line 50 50 Line 50 -50 Line -25} 
\highlight{turquoise!50}{-50} \highlight{limegreen!50}{Ext. 0 0 1 0 0 0 14 0} NewBody \highlight{goldenrod!50}{SOL Circle 0 0 40}
\highlight{goldenrod!50}{SOL Circle 0 0 30} \highlight{coral!50}{Ext. 0 0 1 0 0 0 -10 0} Union
    \end{Verbatim}
    \vspace{-1em}
        \begin{center} \small (b) DeepCAD representation
    \end{center}
    \vspace{-1em}
    \begin{Verbatim}[commandchars=\\\{\}, fontsize=\footnotesize]
import cadquery as cq
w = cq.Workplane('XY') 
w\setlength{\fboxsep}{0pt}\colorbox{turquoise!50}{\strut.box(100,100},\setlength{\fboxsep}{0pt}\colorbox{limegreen!50}{\strut14}).union(
  w\setlength{\fboxsep}{0pt}\colorbox{goldenrod!50}{\strut.sketch().circle(40).circle(30,mode='s')}
  .finalize()\setlength{\fboxsep}{0pt}\colorbox{coral!50}{\strut.extrude(-10)})
    \end{Verbatim}
    \vspace{-1em}
        \begin{center} \small (c) Our CadQuery representation
    \end{center}
        \vspace{-1.5em}
    \caption{Sketch-extrude sequence (top) in DeepCAD representation (middle) and our CadQuery code (bottom).}
\label{fig:formulation}
\end{figure}

Modern feature-based CAD modeling relies on sequences of 2D sketches and operations to create 3D models. Designers first draw geometric primitives (lines, arcs, circles) on a selected plane, then apply operations like extrusion or revolution to generate 3D geometry~\cite{xu2021zonegraph}. As depicted in~\Cref{fig:formulation}(a), we focus on sketch-extrusion sequences, a fundamental CAD modeling pattern widely adopted in previous works~\cite{wu2021deepcad,xu2022skexgen,khan2024cad}. Below, we present our CAD representation, highlighting its advantages over existing language-like encodings, and describe our procedurally generated training data.

\subsection{CadQuery Code} 
Recent approaches in CAD language modeling~\cite{wu2021deepcad, xu2022skexgen, multicad, khan2024cad, dupont2024transcad} encode sketch-extrude sequences as numerical vectors representing features and their parameters as shown in~\Cref{fig:formulation}(b). However, this representation constrains the modeling to specific CAD practices, lacks interpretability, and requires post-processing for CAD kernel compatibility. We propose using CadQuery~\cite{cadquery} Python code to represent sketch-extrude sequences for CAD reverse engineering, offering the following advantages:

\vspace{0.2cm}
\noindent \textbf{Modularity of CAD Features and Design Practices:} Existing language-based CAD reverse engineering methods rely on custom representations of low-level geometric features (lines, arcs, circles) for sketch construction~\cite{wu2021deepcad,seff2022vitruvion}. This approach inherently limits the range of implementable features and design practices. In contrast, CadQuery provides comprehensive built-in CAD functionality, encompassing both low-level features and higher-level geometries like cylinders and boxes as shown in~\Cref{fig:formulation}(c). Furthermore, its programmatic nature enables variable reuse and code modularity. This allows reusing common design features or practices across models, as illustrated by the shared center coordinates across two circles in~\Cref{fig:formulation}~(c). This representation naturally accommodates diverse CAD features and design practices without requiring complex custom encodings or post-processing steps.

\vspace{0.2cm}
\noindent \textbf{Interpretability and LLM Compatibility:} The proposed representation, based on Python and CadQuery syntax, presents an alternative to abstract numerical encodings with improved interpretability. Its code-based format facilitates model editing both programmatically and through CAD software. Importantly, this representation aligns with pre-trained LLMs' demonstrated proficiency in Python code generation and manipulation. Indeed, state-of-the-art proprietary LLMs like GPT-4~\cite{openai2024gpt4technicalreport} achieve over 90\% accuracy on the Python code HumanEval benchmark~\cite{chen2021evaluating}, while even lightweight open-source models such as \mbox{Qwen2-1.5B}~\cite{yang2024qwen2technicalreport} show promising code generation capabilities. Hence, this code representation facilitates fine-tuning of pre-trained LLMs for the specific task of reverse engineering point clouds into CAD Python code and opens the doors for new capabilities with off-the-shelf LLMs. 

\subsection{Procedurally Generated Training Dataset} \label{sec:synth_data}

\begin{figure}[ht]
    \centering
    \includegraphics[width=0.23\linewidth]{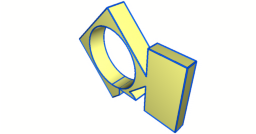}
    \includegraphics[width=0.23\linewidth]{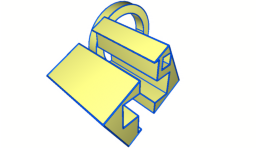}
    \includegraphics[width=0.23\linewidth]{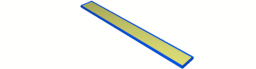}
    \includegraphics[width=0.23\linewidth]{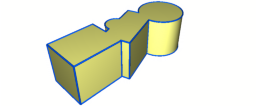}
    \includegraphics[width=0.23\linewidth]{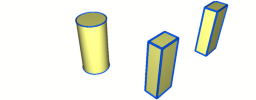}
    \includegraphics[width=0.23\linewidth]{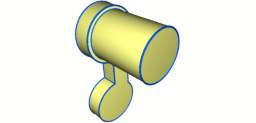}
    \includegraphics[width=0.23\linewidth]{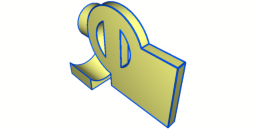}
    \includegraphics[width=0.23\linewidth]{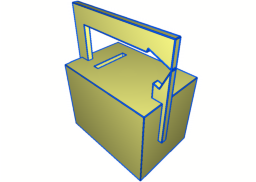}
    \caption{Examples of procedurally generated CAD models.}
    \label{fig:synthetic_data}
\end{figure}

The training of current CAD sketch-extrude reverse engineering methods~\cite{wu2021deepcad, multicad, khan2024cad, dupont2024transcad, ma2024draw} predominantly rely on datasets collected from CAD model repositories~\cite{koch2019abc,cc3d,willis2021fusion}. Considerable efforts are required to parse the CAD models from their original proprietary representations to a suitable one for deep learning~\cite{wu2021deepcad, willis2021fusion}. As a result, existing datasets are restricted not only in scale, but also in control over the design features and patterns included in training.

To address these limitations, we propose to procedurally generate a training dataset of one million CAD models in the form of sketch-extrude sequences written in Python CadQuery~\cite{cadquery} code. Our proposed pipeline randomly generates sketch and CAD operation parameters guided by topological and geometrical heuristics to ensure control over the amount of generated models and the features in the generated codes. The algorithm outlining the steps of this generation pipeline is provided in the supplementary materials along with further statistical analysis of the generated dataset. The modularity of CAD features is incorporated by utilizing both low-level primitives (\ie circles, lines, and arcs) and their abstractions (\ie boxes, cylinders, and rectangles) as well as reusing design elements within the generated sequences. In this work, we focus on some aspects of modularity (\ie, reusing point coordinates, extrusion planes, and abstracting basic shapes such as boxes and cylinders). Further modularity features (\eg, reusing functions corresponding to arbitrary CAD parts, additional CAD operations) can also be integrated in the future. Note that although our generated dataset does not include sequences from human-designed CAD models, it offers significant control over the features and design patterns to which the network is exposed during training. Examples of generated CAD models are shown in~\Cref{fig:synthetic_data}.

\begin{figure*}[!ht]
\begin{center}
\includegraphics[width=.9\textwidth]{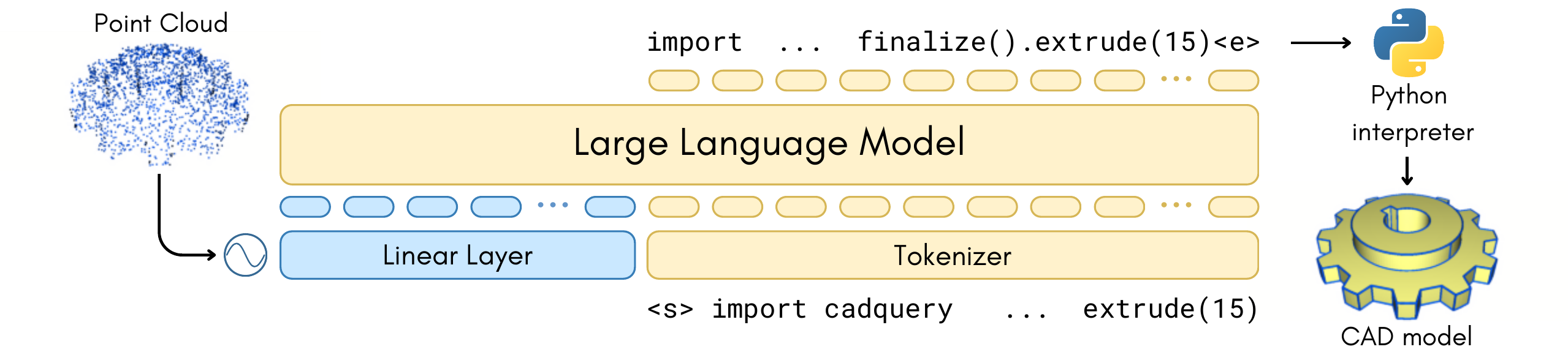}
\end{center}
\setlength{\belowcaptionskip}{-8pt}
\setlength{\abovecaptionskip}{-5pt}
 \caption{Overview of \modelname. The pipeline comprises two parts: (1) a point cloud projector (marked blue) (2) a fine-tuned pre-trained LLM (yellow). An input point cloud is processed using (1), and outputs are then passed to an LLM (2), which predicts a CAD sketch-extrude sequence in the form of executable Python code.}
\label{fig:architecture}
\end{figure*}

\section{\modelname} \label{sec:cad recode}

Building on the introduced CAD code representation and generated training dataset outlined in the previous section, this section introduces \modelname, our proposed model for predicting CAD sketch-extrude sequences as code from input 3D point clouds. We formalize the problem of CAD code prediction, describe the architecture of \modelname, and detail its training and inference processes.

\subsection{Problem Statement}

Let us denote the set of all possible code strings as \mbox{$\Sigma^*$}, where each code string is composed of elements from the finite set of alphanumeric characters and operators in the programming language \mbox{$\Sigma$}. Let \mbox{$\phi_{\text{syn}}: \Sigma^* \rightarrow \{\texttt{True},\texttt{False}\}$} represent the syntactic validation function for Python programming rules (\eg, variable declarations, expression syntax, and statement structure), and \mbox{$\phi_{\text{cad}}: \Sigma^* \rightarrow \{\texttt{True},\texttt{False}\}$} denote the validation function for CAD-specific rules. The latter includes the syntactic validity of the code \wrt to the CAD library, \ie CadQuery~\cite{cadquery}, and the geometric validity of the reconstructed model from the code (\eg, an extrusion can only be applied on a closed loop of 2D primitives, a circle radius cannot be negative). An executable valid CAD code can be formally described by a code string $C \in \mathcal{C}$, where
\[
\mathcal{C} = \{ w \in \Sigma^* \mid \phi_{\text{syn}}(w) \wedge \phi_{\text{cad}}(w) \} \ ,
\]

\noindent represents the set of all valid CAD codes. This formulation ensures that any code string $w$ in $\mathcal{C}$ satisfies both the syntactic requirements of Python ($\phi_{\text{syn}}$) and the CAD code validation rules ($\phi_{\text{cad}}$). Let $\mathbf{P}~=~\{\mathbf{p}_i\}_{i=1}^n~\in~\mathbb{R}^{n \times 3}$ denote an input point cloud, where each point $\mathbf{p}_i~\in~\mathbb{R}^3$ represents 3D Cartesian coordinates. The objective of \modelname~is to learn a mapping

\[
\Psi: \mathbb{R}^{n \times 3} \rightarrow \mathcal{C}, 
\quad C = \Psi(\mathbf{P}) \ , 
\]

\noindent that maps the input point cloud to a valid CAD code $C \in \mathcal{C}$ such that the code, when executed, produces a CAD model geometrically approximating the input point cloud $\mathbf{P}$. Note that the CAD code execution results in a Boundary-Representation (B-Rep)~\cite{lambourne2021brepnet}. Unlike meshes or point clouds, B-Rep is a parametric representation of the CAD model's geometry, enabling seamless integration into modern CAD software and allowing for further modifications. The goal of \modelname~is to infer the CAD code describing the design steps of the CAD model, that when executed results in a B-Rep.

\begin{table*}[!t]
\centering
\resizebox{\textwidth}{!}{
\begin{tabular}{lcccccccccc}
\toprule
\multirow{2}{*}{Method} & \multicolumn{2}{c}{Train Dataset}  & \multicolumn{4}{c}{DeepCAD Test Set} & \multicolumn{4}{c}{Fusion360 Test Set} \\
& Name & Size & Mean CD\textdownarrow & Med. CD\textdownarrow & IoU\textuparrow & IR \textdownarrow & Mean CD\textdownarrow & Med. CD\textdownarrow & IoU\textuparrow & IR\textdownarrow \\
\midrule
DeepCAD~\cite{wu2021deepcad} & DeepCAD & $160$k & $42.5$ & $9.64$ & $46.7$ & $7.1$ & $330$ & $89.2$ & $39.9$ & $25.2$ \\
PrismCAD~\cite{lambourne2022reconstructing} & DeepCAD & $127$k & -- & $4.28$ & $72.1$ & $16.2$ & -- & $4.75$ & $65.3$ & $18.0$ \\
Point2Cyl~\cite{uy2022point2cyl} & DeepCAD & $35$k& -- & $4.27$ & $73.8$ & $3.9$ & -- & $4.18$ & $67.5$ & $3.2$ \\
HNC-CAD~\cite{xu2023hnc} & DeepCAD & $125$k & -- & $8.64$ & $65.3$ & $5.6$ & -- & $36.8$ & $63.5$ & $7.3$ \\
MultiCAD~\cite{multicad} & DeepCAD  & $160$k & -- & $8.09$ & -- & $11.5$ & -- & $42.2$ & -- & $16.5$ \\
TransCAD~\cite{dupont2024transcad} & DeepCAD & $140$k & $32.3$ & $4.51$ & $65.5$ & $1.1$ & $78.6$ & $33.4$ & $60.2$ & $2.4$ \\
CAD-Diffuser~\cite{ma2024draw} & DeepCAD & $160$k & -- & $3.02$ & $74.3$ & $1.5$ & -- & $3.85$ & $63.2$ & $1.7$ \\
CAD-SIGNet~\cite{khan2024cad} & DeepCAD & $160$k & $3.43$ & $0.28$
& $77.6$ & $0.9$ & $7.37$ & $0.48$ & $65.6$ & $1.6$ \\
\textbf{\modelname} & DeepCAD & $160$k & $1.98$ & $0.27$
& $80.7$ & $\mathbf{0.0}$ & $3.37$ & $0.52$ & $67.6$ & $\mathbf{0.1}$ \\
\textbf{\modelname} & Ours & $1$M & $\mathbf{0.30}$ 
& $\mathbf{0.16}$
& $\mathbf{92.0}$ & $0.4$ & $\mathbf{0.35}$
& $\mathbf{0.15}$
& $\mathbf{87.8}$ & $0.5$ \\
\bottomrule
\end{tabular}
}
\caption{Comparison of CAD reverse engineering methods on DeepCAD and Fusion360 datasets. Our \modelname~trained on the $160$\,k DeepCAD dataset demonstrates an improvement over existing state-of-the-art methods both in terms of geometric fidelity and validity of the generated sketch-extrude sequences. Our procedurally generated dataset provides a significant boost in the prediction quality.}
\label{tab:sota}
\end{table*}

\subsection{Proposed Model Architecture}

\modelname~builds on pre-trained LLMs and their prior exposure to Python code, augmenting these with point cloud processing capabilities and CAD-specific Python code knowledge. As shown in~\Cref{fig:architecture}, its architecture consists of two components: (1) a point cloud projector mapping the 3D point cloud into learnable tokens, and (2) a pre-trained LLM-based auto-regressive CAD code decoder. 

\vspace{0.2cm}
\noindent\textbf{Point Cloud Projection Module:}
\modelname~introduces a lightweight projection module $\Psi_p$ that directly maps a dense point cloud $\mathbf{P} \in \mathbb{R}^{n \times d_p}$, where $d_p = 3$ corresponds to the dimension of point coordinates, into a sequence of $n_p~\ll~n$ query tokens $\mathbf{Q}_p = [\mathbf{q}_p^1, \dots, \mathbf{q}_p^{n_p}] \in \mathbb{R}^{n_p \times d_q}$, of embedding dimension $d_q$. The point cloud projector, trained in an end-to-end manner with the CAD code decoder module, consists of three simple components: (1) furthest point sampling to downsample the input point clouds to $n_p$ points, (2) Fourier positional encoding~\cite{zhao2023michelangelo} of coordinates, and (3) a linear layer projecting the encoded coordinates
 into $\mathbf{Q}_p$.

\vspace{0.2cm}
\noindent\textbf{LLM as CAD Code Decoder:}  
Our CAD code decoder, denoted as $\Psi_{\text{LLM}}$, adapts a pre-trained LLM for the specific task of CAD code generation. We leverage the \mbox{Qwen2-1.5B} model~\cite{yang2024qwen2technicalreport} as our LLM backbone, chosen for its balanced trade-off between model capacity and computational requirements. The decoder's input consists of point query tokens $\mathbf{Q}_p$ from the point cloud projector, augmented with $n_t$ code tokens $\mathbf{Q}_t \in \mathbb{R}^{n_t \times d_q}$ obtained by tokenizing the input code as in~\cite{yang2024qwen2technicalreport}. 
The complete input sequence is denoted as $[\mathbf{Q}_p; \mathbf{Q}_t] \in \mathbb{R}^{(n_p + n_t) \times d_q}$, where semicolon indicates concatenation along the sequence dimension. %
The LLM decoder generates the CAD code sequence through next-token prediction. 
As in~\cite{yang2024qwen2technicalreport}, each predicted token 
is mapped to a symbol from the vocabulary $\Sigma$, which includes alphanumeric characters and operators. 

\vspace{0.2cm}
\noindent Overall, \modelname~repurposes the LLM's sequence modeling capabilities for the specialized task of translating geometric point clouds into executable CAD code.

\subsection{Training and Inference Details}
\label{sect:training_inference}

\noindent\textbf{Training Strategy:}
Our training process consists of a single stage. The model operates on query tokens of dimension $d_q = 1536$ and processes input point clouds downsampled to $n_p = 256$ points. Gaussian noise with mean zero and standard deviation of $0.01$ is added to the coordinates of the input points with a probability of $0.5$ per model. The network is trained on the procedurally generated CAD codes, hence exposed to the CAD features and design practices that were included in the algorithm. The training objective minimizes the Negative Log-Likelihood (NLL) of the target CAD code sequence, using special tokens (\textit{$<$s$>$} and \textit{$<$e$>$}) to demarcate sequence boundaries. The point cloud projector $\Psi_p$ learns geometric features from scratch, while the pre-trained decoder $\Psi_{\text{LLM}}$ is fine-tuned for CAD code generation.

\vspace{0.2cm}
\noindent\textbf{Inference Strategy:} At inference time, the point cloud projector $\Psi_p$ processes the input point cloud $\mathbf{P}$ to generate query tokens $\mathbf{Q}_p$, which are then fed to the decoder along with the start token \textit{$<$s$>$}. The model autoregressively generates CAD code tokens until producing a complete code sequence $C$ ending with token \textit{$<$e$>$}. Following~\cite{khan2024cad}, we employ a test-time sampling approach where we generate ten distinct CAD code candidates, each from a different sampling of the input point cloud. For each candidate, we sample points from the predicted CAD model and compute the Chamfer distance \wrt the input point cloud. The candidate with the minimum Chamfer distance is selected as the final output. This verification step effectively favors executable CAD code solutions that are geometrically consistent \wrt the input point cloud.

\section{Experiments} \label{sec:experiments}

In order to validate the effectiveness of \modelname, we conduct a series of experiments across two different scenarios. The first scenario focuses on the reverse engineering task, where the goal is to reconstruct a CAD sketch-extrude sequence in Python code from a given input point cloud. 
The second assesses the interpretability and editibility of the generated CAD code with a proprietary LLM~\cite{openai2024gpt4technicalreport}.

\subsection{Reverse Engineering}

\vspace{0.2cm}
\noindent \textbf{Experimental Setup:} \modelname~is evaluated on three test datasets: DeepCAD~\cite{wu2021deepcad} ($8046$ models), Fusion360~\cite{willis2021fusion} ($1725$ models), and the real-world CC3D~\cite{mallis2023sharp} ($2973$ models). The point clouds are obtained by sampling points on the meshes for DeepCAD and Fusion360. The CC3D dataset provides a real-world scenario with input point clouds sampled from actual 3D scans of CAD models containing surface noise, smoothed edges and missing parts (see supplementary materials for more details). Implementation details are provided in the supplementary. 

\vspace{0.2cm}
\noindent \textbf{Metrics:} 
To evaluate the quality of the predicted CAD sketch-extrude sequences, we use three metrics: Chamfer Distance (CD)~\cite{khan2024cad}, Intersection over Union (IoU)~\cite{ma2024draw}, and Invalidity Ratio (IR)~\cite{wu2021deepcad}. We report both mean and median CD values computed using $8192$ points to assess geometric accuracy. Reported CD values have been multiplied by $10^3$. The IoU is computed from the resulting CAD model meshes and expressed as a percentage. The IR indicates the percentage of generated sequences that fail to produce a valid CAD model.

\begin{figure*}[t]
    \setlength{\tabcolsep}{0pt}
    \centering
    \begin{tabular}{rccccccccc}
    & \multicolumn{3}{c}{DeepCAD Dataset} & \multicolumn{3}{c}{Fusion360 Dataset} & \multicolumn{3}{c}{Real-world CC3D Dataset} \\
    Point Cloud & \includegraphics[width=0.091\linewidth,valign=c]{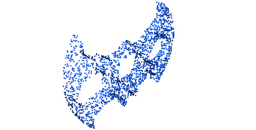} & \includegraphics[width=0.091\linewidth,valign=c]{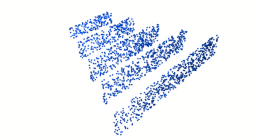} & \includegraphics[width=0.091\linewidth,valign=c]{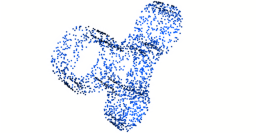} & \includegraphics[width=0.091\linewidth,valign=c]{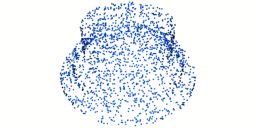} & \includegraphics[width=0.091\linewidth,valign=c]{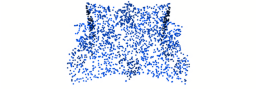} & \includegraphics[width=0.091\linewidth,valign=c]{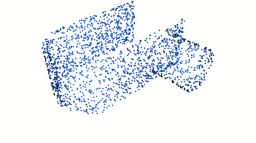} & \includegraphics[width=0.091\linewidth,valign=c]{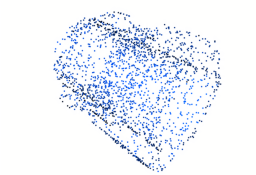} & \includegraphics[width=0.091\linewidth,valign=c]{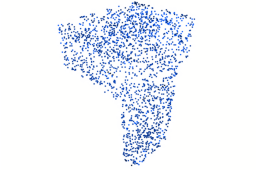} & \includegraphics[width=0.091\linewidth,valign=c]{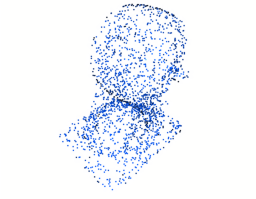} \\
    CAD-SIGNet & \includegraphics[width=0.091\linewidth,valign=c]{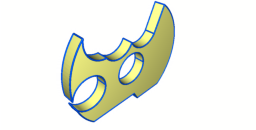} & \includegraphics[width=0.091\linewidth,valign=c]{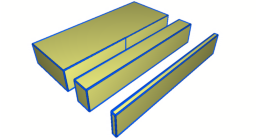} & \includegraphics[width=0.091\linewidth,valign=c]{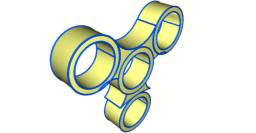} & \includegraphics[width=0.091\linewidth,valign=c]{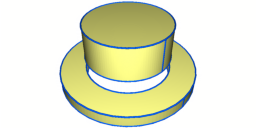} & \includegraphics[width=0.091\linewidth,valign=c]{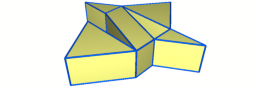} & \includegraphics[width=0.091\linewidth,valign=c]{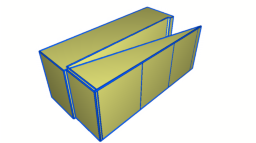} & \includegraphics[width=0.091\linewidth,valign=c]{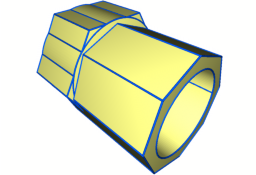} & \includegraphics[width=0.091\linewidth,valign=c]{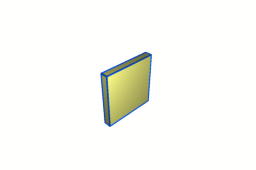} & \includegraphics[width=0.091\linewidth,valign=c]{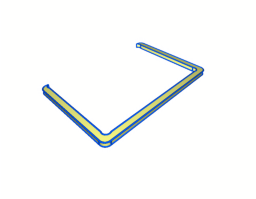} \\
    \modelname & \includegraphics[width=0.091\linewidth,valign=c]{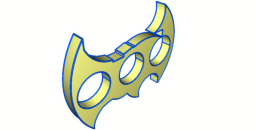} & \includegraphics[width=0.091\linewidth,valign=c]{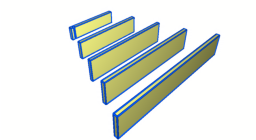} & \includegraphics[width=0.091\linewidth,valign=c]{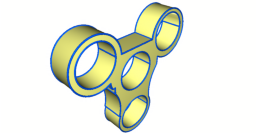} & \includegraphics[width=0.091\linewidth,valign=c]{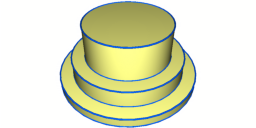} & \includegraphics[width=0.091\linewidth,valign=c]{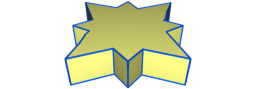} & \includegraphics[width=0.091\linewidth,valign=c]{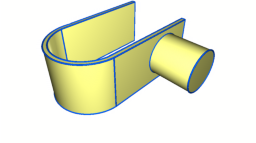} & \includegraphics[width=0.091\linewidth,valign=c]{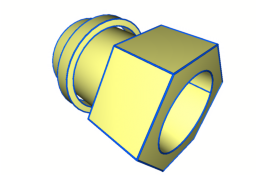} & \includegraphics[width=0.091\linewidth,valign=c]{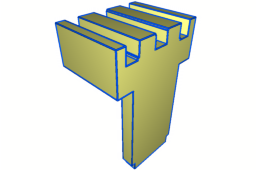} & \includegraphics[width=0.091\linewidth,valign=c]{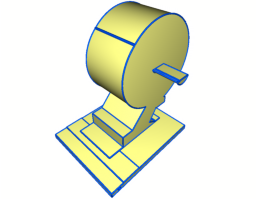} \\
    GT & \includegraphics[width=0.091\linewidth,valign=c]{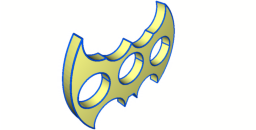} & \includegraphics[width=0.091\linewidth,valign=c]{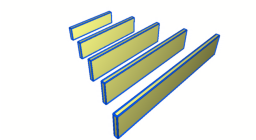} & \includegraphics[width=0.091\linewidth,valign=c]{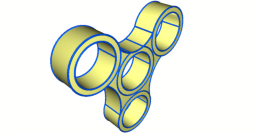} & \includegraphics[width=0.091\linewidth,valign=c]{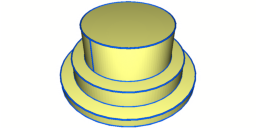} & \includegraphics[width=0.091\linewidth,valign=c]{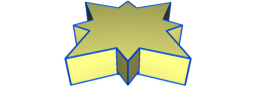} & \includegraphics[width=0.091\linewidth,valign=c]{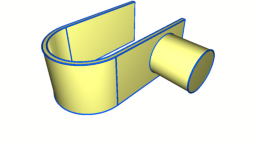} & \includegraphics[width=0.091\linewidth,valign=c]{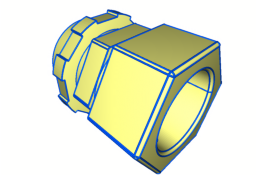} & \includegraphics[width=0.091\linewidth,valign=c]{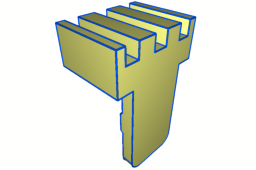} & \includegraphics[width=0.091\linewidth,valign=c]{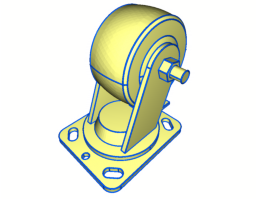} \\
    \end{tabular}
    \setlength{\belowcaptionskip}{-8pt}
    \caption{Qualitative results on the DeepCAD, Fusion360, and CC3D datasets. For each input point cloud (first row), we compare CAD models produced by CAD-SIGNet (second) and our \modelname~trained on our dataset (third) with a ground truth CAD model (bottom row). While CAD-SIGNet often fails to restore the general shape, CAD-Recode outputs only slightly deviate from ground truth in most cases.}
    \label{fig:example_images}
\end{figure*}

\vspace{0.2cm} 
\noindent \textbf{Results \& Analysis:} \Cref{tab:sota} presents results on the test sets of DeepCAD and Fusion360 datasets, where \modelname~establishes new state-of-the-art performance across all metrics. Note that the results of state-of-the-art methods in~\Cref{tab:sota} are borrowed from~\cite{ma2024draw}, except for CAD-SIGNet~\cite{khan2024cad}, MultiCAD~\cite{multicad}, TransCAD~\cite{dupont2024transcad}, and DeepCAD~\cite{wu2021deepcad} which were taken from~\cite{khan2024cad} and~\cite{dupont2024transcad}. First, we convert the DeepCAD dataset ($160\,\text{k}$ models) to CadQuery Python code and use it to train \modelname~(results are in row before last of~\Cref{tab:sota}). When trained on DeepCAD dataset as existing methods, \modelname~outperforms them in almost all metrics. These results showcase the effectiveness of \modelname~and the proposed CAD code representation. 

Training on $1\,\text{M}$ generated samples results in substantial improvements in CD and IoU metrics while maintaining a negligibly low invalidity ratio (last row of~\Cref{tab:sota}), reflecting significantly better geometric fidelity in the predicted CAD models. \modelname~demonstrates a ten-fold improvement in mean CD and an increase of IoU by over $10\%$ on both DeepCAD and Fusion360 datasets compared to the existing best methods. These results confirm that our large-scale procedurally generated training dataset provides substantial benefits.

As illustrated in~\Cref{fig:example_images}, this translates to consistent reconstruction quality, where \modelname~reliably produces CAD models that accurately capture the geometry from the input point cloud. In contrast, CAD-SIGNet~\cite{khan2024cad} can generate shapes that deviate significantly from the target geometry, further highlighting the advantages of our approach.

\begin{table}
\centering
\setlength{\tabcolsep}{3pt}
\begin{tabular}{lcccc}
\toprule
Method & Mean CD\textdownarrow & Med. CD\textdownarrow & IoU\textuparrow & IR\textdownarrow \\
\midrule
DeepCAD~\cite{wu2021deepcad} & -- & $263$ & -- & $12.7$ \\
CAD-SIGNet~\cite{khan2024cad} & $14.82$ & $2.90$ & $42.6$ & $2.5$ \\
\textbf{\modelname} & $\mathbf{0.76}$
& $\mathbf{0.31}$
& $\mathbf{74.2}$ & $\mathbf{0.3}$ \\
\bottomrule
\end{tabular}%
\setlength{\belowcaptionskip}{-8pt}
\caption{Results on the CC3D dataset, where input point clouds are sampled from real 3D scans. \modelname~ significantly outperforms DeepCAD, and CAD-SIGNet.}
\label{tab:cc3d}
\end{table}

\noindent\textbf{Real-world Scenario:} In~\Cref{tab:cc3d}, we evaluate \modelname~on the real-world CC3D dataset, where input point clouds are sampled from 3D scans and contain artifacts such as surface noise, smooth edges, and missing parts. Even under these challenging conditions, our method achieves significant improvements over CAD-SIGNet~\cite{khan2024cad}, with a $89\%$ lower median CD and a $30\%$ higher IoU, while maintaining a low IR. From the CC3D qualitative results in~\Cref{fig:example_images}, \modelname~is able to recover geometries that are much closer to the ground truth than current state-of-the-art. However, it can be observed that \modelname~still lacks the expressiveness to model complex shapes that contain operations beyond the extrusion operation such as revolution and fillet. This can be attributed to the choice of features and design practices in the procedurally generated training dataset. Nevertheless, we believe that this can be addressed in future works by incorporating further features in the dataset generation procedure. Our results on CC3D are compared with methods previously reported for this dataset~\cite{khan2024cad}, namely CAD-SIGNet and DeepCAD.

\vspace{0.2cm}
\noindent \textbf{Ablation Study:}
To evaluate the different components of \modelname, we conduct a comprehensive ablation study on the amount of training data, test-time sampling, and the number of input points and model parameters.

\begin{table*}[t]
\centering
\resizebox{\textwidth}{!}{%
\begin{tabular}{lcccccccccccc}
\toprule
\multirow{2}{*}{Method} & \multicolumn{2}{c}{Train Dataset} & Test-time & \multicolumn{3}{c}{DeepCAD} & \multicolumn{3}{c}{Fusion360} & \multicolumn{3}{c}{Real-World CC3D} \\
& Name & Size & Sampling & CD\textdownarrow & IoU\textuparrow & IR\textdownarrow & CD\textdownarrow & IoU\textuparrow & IR\textdownarrow & CD\textdownarrow & IoU\textuparrow & IR\textdownarrow \\
\midrule
Previous best~\cite{khan2024cad} & DeepCAD & $160$\,k & \cmark & $3.43$ & $77.6$ & $0.9$ & $7.37$ & $65.6$ & $1.6$ & $14.80$ & $42.6$ & $4.4$ \\
\modelname & DeepCAD & $160$\,k & \cmark & $1.98$ & $80.7$ &  $0.0$ & $3.37$ & $67.6$ & $0.1$ & $3.79$ & $56.4$ & $0.0$ \\
\modelname & Ours & $160$\,k & \cmark & $0.54$
& $88.3$ & $0.3$ & $0.66$
& $82.0$ & $0.1$ & $1.27$ & $69.0$ & $0.2$ \\
\modelname & Ours & $1$\,M& \cmark & $\mathbf{0.30}$
& $\mathbf{92.0}$ & $0.4$ & $\mathbf{0.35}$
& $\mathbf{87.8}$ & $0.5$ & $\mathbf{0.76}$
& $\mathbf{74.2}$ & $0.3$ \\
\midrule
Previous best~\cite{khan2024cad} & DeepCAD & $160$\,k & \xmark & $6.81$ & $77.3$ & $4.4$ & $14.5$ & $58.4$ & $9.3$ & $32.59$ & $39.1$ & $15.5$ \\
\modelname & Ours & $1$\,M& \xmark & $\mathbf{0.75}$ & $\mathbf{89.3}$ & $4.9$ & $\mathbf{0.89}$ & $\mathbf{84.2}$ & $8.7$ & $\mathbf{3.05}$ & $\mathbf{65.6}$ & 16.8 \\
\bottomrule
\end{tabular}%
}
\caption{Ablation of training data and test-time sampling. The results demonstrate the advantage of training on our procedurally generated data, while the test-time sampling helps reducing the invalidity ratio. CD stands for mean Chamfer distance.}
\label{tab:ablation}
\end{table*}

Training \modelname~on $160$\,k procedurally generated samples using the method described in~\Cref{sec:synth_data} leads to significant improvements in geometric fidelity of the predicted samples over training on the DeepCAD dataset with the same amount of data (see row 2 and 3 of~\Cref{tab:ablation}). Furthermore, scaling our training dataset to $1$\,M samples provides further improvements across all datasets (row 4 of~\Cref{tab:ablation}). As compared to DeepCAD training dataset, our procedural dataset generation provides a better way of learning the mapping between point clouds and CAD codes which can be further improved by scaling up the dataset size. 

We investigate the effectiveness of the test-time sampling approach that generates multiple CAD code predictions through different point cloud samplings, as described in~\Cref{sect:training_inference}. As shown in the third and last row of~\Cref{tab:ablation}, the test-time sampling approach mainly helps reducing the ratio of invalid predicted CAD codes (IR). For comparison, CAD-SIGNet~\cite{khan2024cad} employs a probability-based sampling. Yet, even without test-time sampling our method still performs better on the reconstruction metrics than CAD-SIGNet~\cite{khan2024cad}. 

\begin{table}
\centering
\resizebox{\linewidth}{!}{
\begin{tabular}{cccccccc}
\toprule
\multirow{2}{*}{Points} & Model & \multicolumn{2}{c}{DeepCAD} & \multicolumn{2}{c}{Fusion360} & \multicolumn{2}{c}{CC3D} \\
& Size & CD\textdownarrow & IoU\textuparrow & CD\textdownarrow & IoU\textuparrow & CD\textdownarrow & IoU\textuparrow \\
\midrule
$128$ & $0.5$\,B & $0.18$ & $89.9$ & $0.18$ & $84.3$ & $0.38$ & $71.9$ \\
$256$ & $0.5$\,B & $0.17$ & $90.6$ & $0.17$ & $85.4$ & $0.36$ & $72.6$ \\
$256$ & $1.5$\,B & $\mathbf{0.16}$ & $\mathbf{92.0}$ & $\mathbf{0.15}$ & $\mathbf{87.8}$ & $\mathbf{0.31}$ & $\mathbf{74.2}$ \\
\bottomrule
\end{tabular}}
\caption{Ablation of architecture details. CD stands for median Chamfer distance.}
\label{tab:parameters}
\end{table}

Results in~\Cref{tab:parameters} show an ablation of the input point cloud size and the number of parameters of the LLM backbone. It can be observed that using an input point cloud of $256$ points and Qwen1.5b results in the highest IoU. This setting with a relatively small input point cloud and lightweight LLM backbone provides the best balance between prediction accuracy and memory requirements. Results on all metrics are included in the supplementary materials. 

\subsection{CAD-QA and Editability}
\label{sect:interpret_edit}

\noindent \textbf{CAD-QA and LLM Interpretability:} CAD SGP-Bench~\cite{qiu2024can} is a benchmark of $1000$ CAD-specific Question Answering (CAD-QA) tasks that test LLMs' understanding of CAD model geometry from sketch-extrude sequences encoded as in DeepCAD~\cite{wu2021deepcad}. These questions require analyzing various geometric aspects, such as relative sizes and 3D primitive types. We extend this benchmark to evaluate CAD reverse engineering methods by using point clouds as input instead of CAD sequences. Our evaluation follows a two-stage process: first predicting sketch-extrude sequences from point clouds as CadQuery code with \modelname, then using GPT-4o~\cite{openai2024gpt4technicalreport} to answer CAD-specific questions. Without requiring additional interpretation hints, our approach achieves $76.5\%$ accuracy on this CAD-QA task (\Cref{tab:vqa}). For comparison, we evaluate two baseline approaches: CAD-SIGNet~\cite{khan2024cad} and PointLLM~\cite{xu2024pointllm}. When using CAD-SIGNet's output with GPT-4o, even with provided interpretation hints explaining the sequence format, the accuracy reaches only $63.2\%$. PointLLM, which directly processes point clouds for language tasks, achieves $42.3\%$ accuracy when prompted with the CAD-specific questions. These results demonstrate that \modelname~effectively captures CAD geometric information while generating an output in a format that proprietary LLMs can naturally interpret and process.

\begin{table}
\centering
\setlength{\tabcolsep}{5pt}
\begin{tabular}{lc}
\toprule
Method & CAD-QA Accuracy\textuparrow \\ 
\midrule
PointLLM~\cite{xu2024pointllm} & $42.3$ \\
CAD-SIGNet~\cite{khan2024cad} \textrightarrow\ GPT-4o & $63.2$ \\
\textbf{\modelname} \textrightarrow\ GPT-4o & $\mathbf{76.5}$ \\
\bottomrule
\end{tabular}
\caption{Point cloud CAD-specific question answering (CAD-QA)
on the SGP-Bench benchmark. Our \modelname~supplied with a GPT-4o significantly outperforms baseline methods.}
\label{tab:vqa}
\end{table}

\begin{figure}[ht]
\begin{center}
\includegraphics[width=\linewidth]{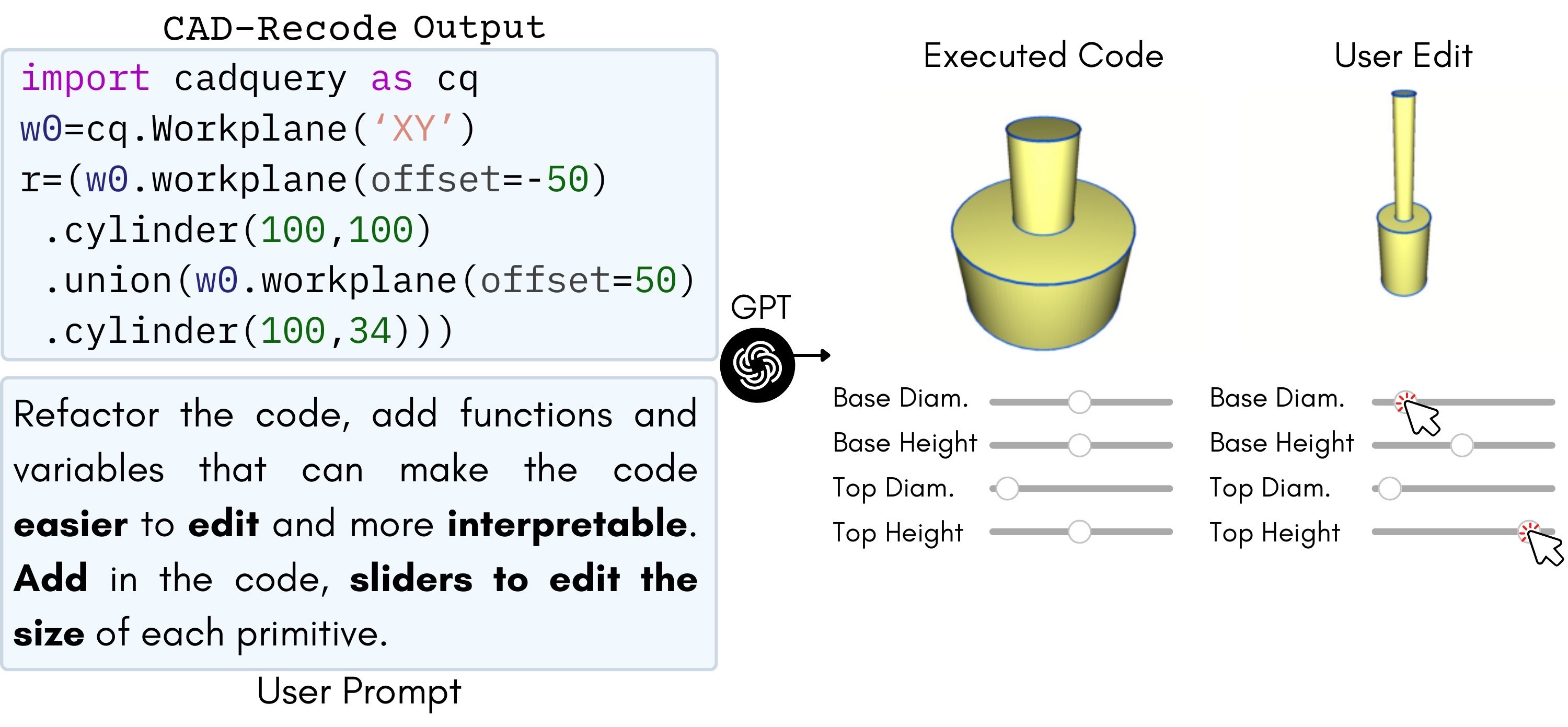}
\end{center}
\setlength{\abovecaptionskip}{-5pt}
\caption{Interactive editing of a CAD model. Given the code output from \modelname~and a generic prompt, GPT-4o allows automated and interactive editing of the CAD model.}
\label{fig:editing}
\end{figure}

\vspace{0.2cm}
\noindent \textbf{Editing Pipeline:} Leveraging the interpretable nature of our code-based output, we present an automated editing pipeline using GPT-4o~\cite{openai2024gpt4technicalreport}. Through a simple prompt, the system refactors the generated code to expose geometric parameters via interactive sliders, enabling direct manipulation of the reconstructed model. As shown in~\Cref{fig:editing}, the resulting code can be directly executed in a Python environment to provide an interactive editing interface. Implementation details are provided in the supplementary materials.

\section{Conclusion} \label{sec:conclusion}

This work rethinks the problem of feature-based CAD reverse engineering by approaching it through pre-trained LLMs taking advantage of CAD Python-based representation. Our key contributions include a new CAD code representation for reverse engineering sketch-extrude sequences, very large-scale procedurally generated training dataset in the form of CAD programs, and a point cloud-to-CAD code model. We demonstrate that \modelname~outperforms existing methods by a large margin on three datasets, including the real-world CC3D dataset. We also showcase that combining pre-trained LLMs with geometric understanding enables powerful new workflows, where designers can reconstruct CAD models from point clouds and modify them through natural language. We believe that this work will open new perspectives for CAD reverse engineering. We identify the following interesting future works: (1) further exploiting the modularity of the proposed CAD code representation, (2) scaling up the LLM and the dataset to enable reverse engineering of more complex CAD models.

\vspace{0.2cm}
\noindent \textbf{Acknowledgement:} The present project is supported by the National Research Fund, Luxembourg under the BRIDGES2021/IS/16849599/FREE-3D and IF/17052459/CASCADES projects, and by Artec 3D.

\section*{Appendix}

\appendix

\section{Training Details}

The \modelname~implementation uses Qwen2-1.5B as the LLM decoder. The training configuration employs the AdamW optimizer with a learning rate of $0.0002$ and weight decay of $0.01$, while maintaining other parameters at their default values from the HuggingFace implementation~\cite{hugging}, including the cosine learning rate scheduler. The training process is conducted for $100\, \text{k}$ iterations, incorporating an initial warmup period of $1\, \text{k}$ iterations. Using a single NVIDIA H100 GPU with a batch size of 18, the complete training process takes approximately $12$ hours. For ablation study examining decoder size impact (Section~\textcolor{cvprblue}{5.1} of the main paper), we utilize Qwen2-0.5B.

\section{Training Dataset Generation Algorithm}

In Section~\textcolor{cvprblue}{3.2}, the procedurally generated training dataset is presented. The main advantage of generating data over using the existing DeepCAD dataset for training is that the algorithm allows full control over the amount of data as well as the features and design patterns that the network is exposed to during training. We generate one million valid Python CadQuery code snippets, through an automated pipeline leveraging PythonOCC~\cite{opencascade} and CadQuery~\cite{cadquery}. The generation process consists of two primary components: (1) a sketch profile generator (\Cref{alg:sketch-generation}) that creates valid 2D sketches, and (2) a CAD model generator (\Cref{alg:cad-generation}) that produces 3D CAD models from these sketches.

The sketch generation process combines primitive shapes (circles and rectangles) through boolean operations (union and cut). From each generated sketch, we extract the primitives (lines, arcs, and circles) from both inner and outer loops. The validity of the generated sketch is ensured through multiple verification steps, including verifying that loops do not intersect, and each primitive has a length greater than zero. Finally, we ensure that the randomly generated CAD code has not previously been generated using the duplicate detection protocol outlined in~\cite{xu2022skexgen}. This ensures that each sample in the dataset is unique.

The CAD model generation procedure extrudes the validated sketches and combines them through union operations. The planes on which the sketches lie are randomly generated by choosing one of the three canonical planes translated by a random amount. Each resulting model undergoes normalization to fit within a unit bounding box centered at the origin. The parameters are quantized so that the coordinates of any point on the CAD surface are within the range $-100$ to $100$ with a minimum resolution of $1$ unit. We then simplify the sequence using higher level abstractions (rectangle, box, and cylinder) by considering the sequence parameters. Our validation framework verifies that a generated code $w$ executes without errors ($\phi_{\text{syn}}$). Furthermore, we check that the executed code produces a geometric valid CAD model ($\phi_{\text{cad}}$) using the \texttt{BRepCheck\_Analyzer} function from PythonOCC as in~\cite{wu2021deepcad}. Invalid models are excluded from the dataset.

\begin{algorithm*}
\footnotesize
\caption{Generate2DSketch}
\label{alg:sketch-generation}
\begin{algorithmic}[1]
\Function{Generate2DSketch}{}
    \State $numPrimitives \gets RandInt(3, 8)$ \Comment{Choose random number of shape primitives}
    \State $compositeShape \gets \emptyset$ \Comment{Initialize empty shape}
    
    \For{$i \gets 1$ to $numPrimitives$} \Comment{Build shape by combining primitives}
        \State $primitive \gets$ random from \{Circle, RotatedRectangle\}
        \State $booleanOperation \gets$ random from \{Union, Cut\} \Comment{Union adds, Cut subtracts}
        \State $compositeShape \gets ApplyOperation(compositeShape, primitive, booleanOperation)$
    \EndFor
    
    \State $boundaryLoops \gets ExtractBoundaryLoops(compositeShape)$ \Comment{Extract shape boundaries}
    \State $boundaryComponents \gets \emptyset$
    
    \For{$loop \in boundaryLoops$} \Comment{Process each boundary loop}
        \State $(edgeSequence, isOuter) \gets AnalyzeBoundary(loop)$ 
        \Comment{List of parametric curves (lines, arcs, circles)}
        \State $boundaryComponents.Append((edgeSequence, isOuter))$
    \EndFor
    
    \State $boundaryComponents \gets ValidateShapeTopology(boundaryComponents)$ \Comment{Ensure valid shape topology}
    \State \Return $boundaryComponents$ \Comment{Returns list of (edges, boolean) tuples}
\EndFunction
\end{algorithmic}
\end{algorithm*}

\begin{algorithm*}
\footnotesize
\caption{GenerateCAD}
\label{alg:cad-generation}
\begin{algorithmic}[1]
\Function{GenerateCAD}{}
    \State $cadModel \gets \emptyset$ \Comment{Initialize empty CAD model}
    \State $planes \gets GenerateRandomPlanes()$ \Comment{Create set of reference planes}
    \State $sketches \gets Generate2DSketch()$ \Comment{Get sketches from Algorithm~\ref{alg:sketch-generation}}
    
    \For{$sketch \in sketches$} \Comment{Create 3D volumes from sketches}
        \State $plane \gets RandomSelect(planes)$ \Comment{Select random reference plane}
        \State $volume \gets ExtrudeSketch(sketch, plane)$ \Comment{Create 3D volume by extrusion}
        \State $cadModel \gets BooleanUnion(cadModel, volume)$ \Comment{Add volume to model}
    \EndFor
    
    \State $cadModel \gets NormalizeModel(cadModel)$ \Comment{Ensure the model fits within a unit box}
    \State $cadModel \gets QuantizeParameters(cadModel)$ \Comment{Discretize model parameters}
        \State $cadModel \gets SimplifyCADModel(cadModel)$ \Comment{Identify high-level abstractions (rectangle, box, and cylinder)}
    \State $cadModel \gets ValidateCADModel(cadModel)$ \Comment{Ensure validity of CadQuery code and CAD model geometry}
    \State $cadModel \gets CheckDuplicate(cadModel)$ \Comment{Ensure that the sequence has not previously been generated.}
    \State \Return $cadModel$
\EndFunction
\end{algorithmic}
\end{algorithm*}

\begin{figure*}
\begin{tabular}{cc}
\includegraphics[width=0.091\linewidth,valign=c]{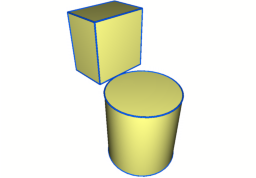} &
\begin{minipage}[c]{0.85\linewidth}
\begin{pythoncode}
import cadquery as cq
w0 = cq.Workplane('ZX', origin=(0, -13, 0))
r = w0.workplane(offset=-87 / 2).moveTo(52.5, 10.5).box(57, 83, 87)
  .union(w0.workplane(offset=23 / 2).moveTo(-29, 0).cylinder(23, 30))
  .union(w0.workplane(offset=113 / 2).moveTo(-29, 0).cylinder(113, 52))
\end{pythoncode}
\end{minipage} \\ \\ 
\includegraphics[width=0.091\linewidth,valign=c]{images/render_iccv/train/519020.png} &
\begin{minipage}[c]{0.85\linewidth}
\begin{pythoncode}
import cadquery as cq
w0 = cq.Workplane('ZX', origin=(0, -30, 0))
r = w0.sketch().segment((-30, -27),(-18, -31)).segment((-19, -31)).segment((-19, -100))
    .segment((38, -100)).segment((38, -31)).segment((10, -31)).segment((13, -23))
    .arc((30, -13), (23, 5)).segment((33, 33)).segment((16, 39)).arc((-12, 99),(-9, 33))
    .close().assemble().finalize().extrude(60)
\end{pythoncode}
\end{minipage} \\ \\
\includegraphics[width=0.091\linewidth,valign=c]{images/render_iccv/train/196160.png} &
\begin{minipage}[c]{0.85\linewidth}
\begin{pythoncode}
import cadquery as cq
w0 = cq.Workplane('YZ', origin=(-14, 0, 0))
r = w0.workplane(offset=17 / 2).moveTo(4, -73.5).box(104, 53, 17)
  .union(w0.sketch().segment((-78, 23), (2, -55)).segment((40, -17))
    .arc((42, -24),(48, -30)).segment((48, 5)).segment((61, 5))
    .segment((78, 22)).segment((-2, 100)).close().assemble()
    .push([(0, 22)]).circle(50, mode='s').finalize().extrude(29))
\end{pythoncode}
\end{minipage} \\ \\
\includegraphics[width=0.091\linewidth,valign=c]{images/render_iccv/train/339060.png} &
\begin{minipage}[c]{0.85\linewidth}
\begin{pythoncode}
import cadquery as cq
w0 = cq.Workplane('XY', origin=(0, 0, 42))
w1 = cq.Workplane('YZ', origin=(-17, 0, 0))
r = w0.sketch().arc((-12, 6), (34, -29),(-1, 16)).segment((5, 4)).segment((-8, -2))
    .close().assemble().finalize().extrude(56)
  .union(w0.sketch().arc((-42, 54), (-12, 71), (19, 54)).segment((19, 78))
    .segment((-42, 78)).close().assemble().finalize().extrude(58))
  .union(w1.sketch().segment((-44, -100), (51, -100)).segment((51, 5)).segment((27, 5))
    .arc((-58, 40),(-44, -51)).close().assemble().reset()
    .face(w1.sketch().arc((-54, -17), (-26, -34),(3, -17)).close()
    .assemble(), mode='s').reset().face(w1.sketch().segment((-54, 14), (3, 14))
    .arc((-26, 31), (-54, 14)).assemble(), mode='s').finalize().extrude(-13))
\end{pythoncode}
\end{minipage} \\ \\
\includegraphics[width=0.091\linewidth,valign=c]{images/render_iccv/train/243730.png} &
\begin{minipage}[c]{0.85\linewidth}
\begin{pythoncode}
import cadquery as cq
w0 = cq.Workplane('YZ', origin=(-22, 0, 0))
w1 = cq.Workplane('ZX', origin=(0, -19, 0))
r = w0.sketch().segment((-100, -83),(-67, -83)).segment((-80, -52)).segment((-75, -50))
    .segment((-75, 62)).segment((17, 62)).segment((17, -62)).segment((-40, -62))
    .segment((-37, -71)).segment((-65, -83)).segment((43, -83))
    .segment((43,83)).segment((-100,83)).close().assemble().finalize().extrude(8)
  .union(w1.sketch().segment((-77, -53),(76, -53)).arc((76, -48),(77, -42))
  .segment((77, 53)).segment((-77, 53)).close().assemble()
  .push([(38.5, 2.5)]).rect(9, 57, mode='s').finalize().extrude(119))
\end{pythoncode}
\end{minipage}
\end{tabular}
\caption{Examples from our procedurally generated training dataset. Each row contains CadQuery Python code and a corresponding CAD model. Examples contain not only basic \textit{line}, \textit{circle}, and \textit{arc} primitives, but also higher-level abstractions such as \textit{rect}, \textit{box}, and \textit{cylinder}.}
\label{fig:train_with_code}
\end{figure*}

\Cref{fig:train_with_code} presents examples of CAD models alongside their corresponding CadQuery Python code from our procedurally generated dataset. It is worth noting that the generated codes are fairly compact, this was designed to facilitate training. All code examples are directly executable using a standard Python interpreter with the CadQuery library. The codes follow a consistent three-part structure: (1) necessary library import, (2) definition of sketch planes, and (3) sketch-extrude operations combined through union. 

\begin{figure*}
    \centering
    \begin{subfigure}[b]{0.48\textwidth}
        \centering
        \includegraphics[width=\linewidth]{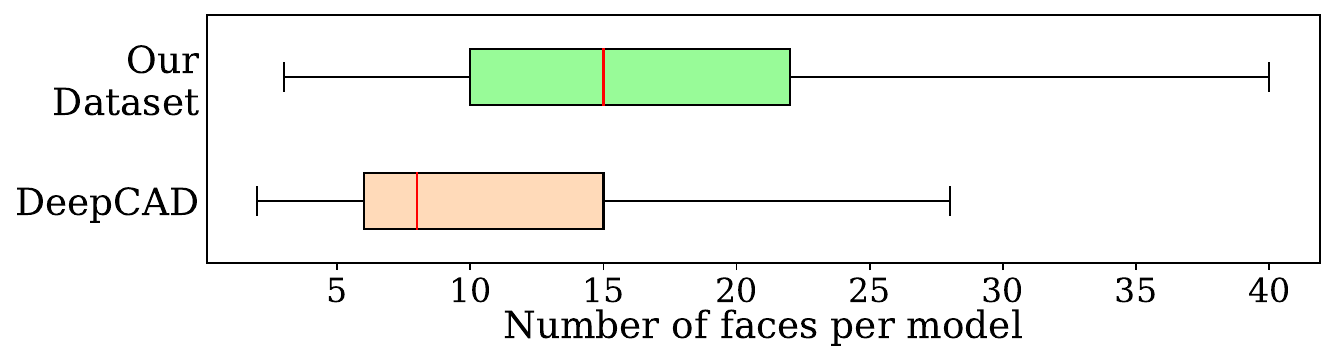}
        \caption{Box-plot graph of the distribution of the number of faces per model.}
        \label{fig:boxplot_faces}
    \end{subfigure}%
    \hfill
    \begin{subfigure}[b]{0.48\textwidth}
        \centering
        \includegraphics[width=\linewidth]{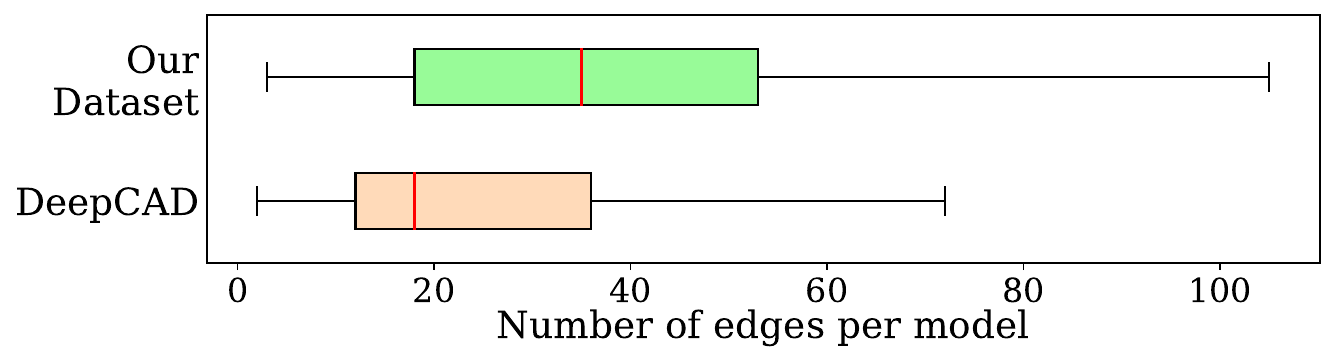}
        \caption{Box-plot graph of the distribution of the number of edges per model.}
        \label{fig:boxplot_edges}
    \end{subfigure}
    \setlength{\belowcaptionskip}{-4pt}
    
    \caption{Our $1$\,M procedurally generated training dataset displays distributions CAD models that are skewed towards models with larger edge and face count per model than the DeepCAD dataset ($160$\,k models).}
    \label{fig:boxplots}
\end{figure*}

The training dataset generation procedure provides full control over the features included. In~\Cref{fig:boxplots}, it can be observed that the distribution of our CAD models is skewed towards models with larger face and edge count per model with interquartile ranges. As a result, our procedurally generated dataset provides a larger variety of models.

\section{Real-World CC3D Dataset}
\begin{figure*}[htb!]
    \setlength{\tabcolsep}{0pt}
    \centering
    \begin{tabular}{rccc}
    \shortstack{Ground truth \\ CAD model} & \includegraphics[width=0.27\linewidth,valign=c]{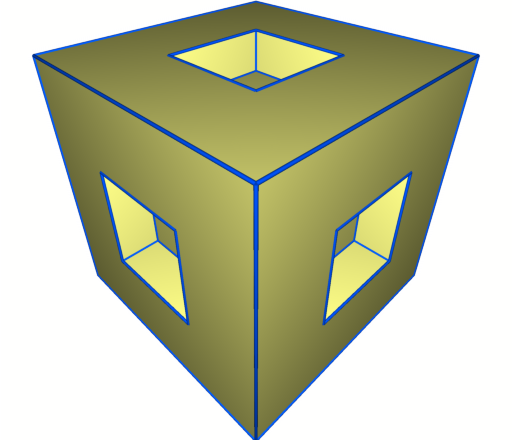} & \includegraphics[width=0.27\linewidth,valign=c]{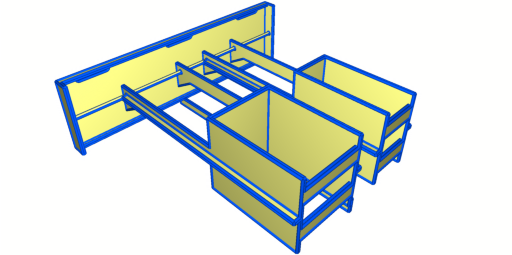} & \includegraphics[width=0.27\linewidth,valign=c]{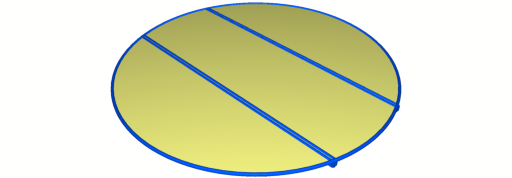} \\
    \shortstack{Real-world \\ noisy scan} & \includegraphics[width=0.27\linewidth,valign=c]{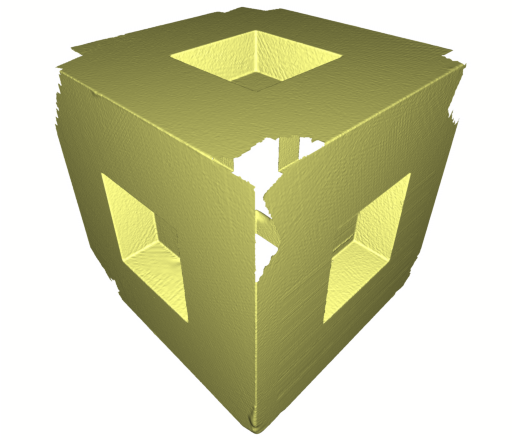} & \includegraphics[width=0.27\linewidth,valign=c]{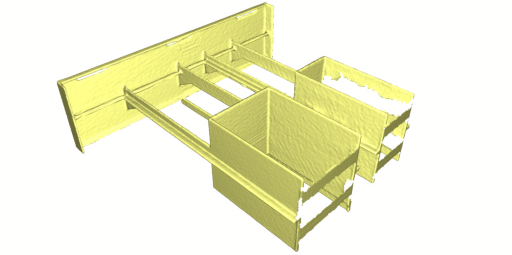} & \includegraphics[width=0.27\linewidth,valign=c]{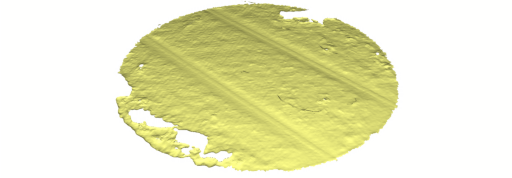} \\
    \end{tabular}
\caption{Example models from real-world CC3D dataset. The scans exhibits numerous artifacts such as surface noise, missing parts and smoothed edges. In the CC3D experiments reported in the main paper, the input point clouds are sampled from the scans. Zoom in for better details.}
\label{fig:cc3d_scans}
\end{figure*}

Results on the real-world CC3D~\cite{cc3d, mallis2023sharp} dataset are presented in Table~\textcolor{cvprblue}{3} of the main paper. This scenario provides an experimental evaluation in a realistic setting, as the input point clouds are sampled from actual 3D scans of CAD models. Sample models are depicted in~\Cref{fig:cc3d_scans}, where artifacts such as surface noise, smoothed edges, and missing parts can be observed. Furthermore, several models from the CC3D dataset are constructed using a range of operations beyond simple extrusion, including revolution, chamfer, and fillet. Consequently, the real-world CC3D dataset provides a challenging set of inputs that enables robust in-the-wild evaluation of our proposed method.

\section{Further Experimental Results}

\begin{figure*}[!htb]
    \setlength{\tabcolsep}{0pt}
    \centering
    \begin{tabular}{rccccccccc}
    Point Cloud & \includegraphics[width=0.091\linewidth,valign=c]{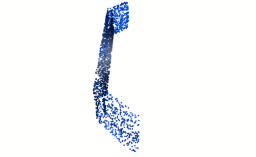} & \includegraphics[width=0.091\linewidth,valign=c]{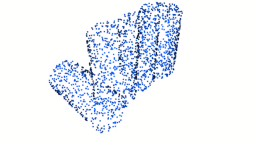} & \includegraphics[width=0.091\linewidth,valign=c]{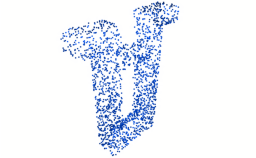} & \includegraphics[width=0.091\linewidth,valign=c]{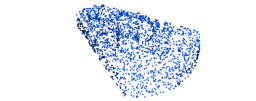} & \includegraphics[width=0.091\linewidth,valign=c]{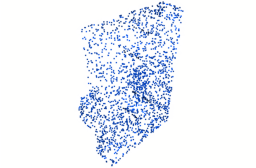} & \includegraphics[width=0.091\linewidth,valign=c]{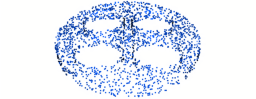} & \includegraphics[width=0.091\linewidth,valign=c]{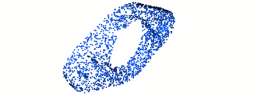} & \includegraphics[width=0.091\linewidth,valign=c]{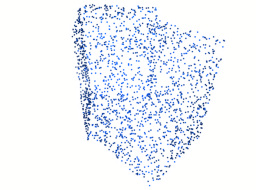} & \includegraphics[width=0.091\linewidth,valign=c]{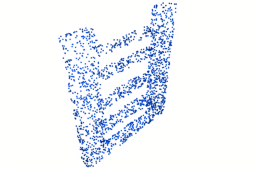} \\
    CAD-SIGNet & \includegraphics[width=0.091\linewidth,valign=c]{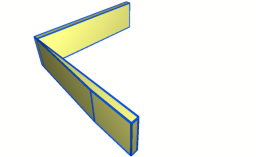} & \includegraphics[width=0.091\linewidth,valign=c]{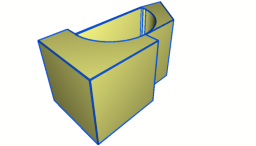} & \includegraphics[width=0.091\linewidth,valign=c]{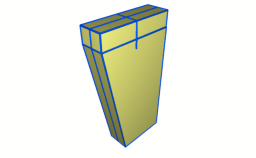} & \includegraphics[width=0.091\linewidth,valign=c]{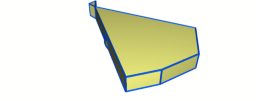} & \includegraphics[width=0.091\linewidth,valign=c]{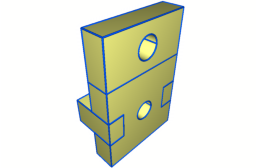} &  \includegraphics[width=0.091\linewidth,valign=c]{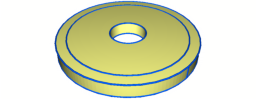} & \includegraphics[width=0.091\linewidth,valign=c]{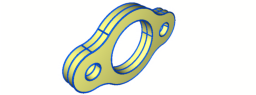} & \includegraphics[width=0.091\linewidth,valign=c]{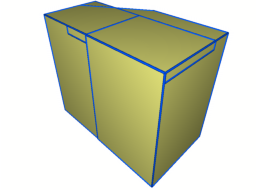} & \includegraphics[width=0.091\linewidth,valign=c]{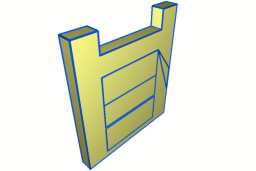} \\
    \modelname & \includegraphics[width=0.091\linewidth,valign=c]{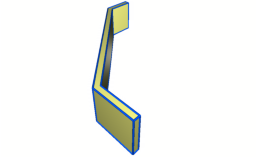} & \includegraphics[width=0.091\linewidth,valign=c]{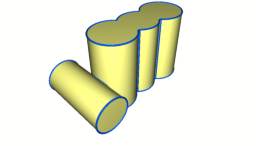} & \includegraphics[width=0.091\linewidth,valign=c]{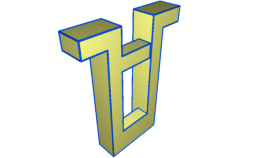} & \includegraphics[width=0.091\linewidth,valign=c]{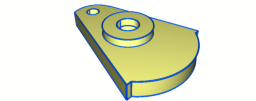} & \includegraphics[width=0.091\linewidth,valign=c]{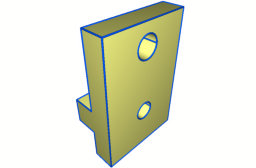} &  \includegraphics[width=0.091\linewidth,valign=c]{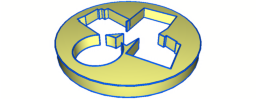} & \includegraphics[width=0.091\linewidth,valign=c]{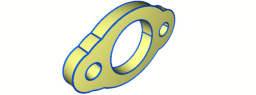} & \includegraphics[width=0.091\linewidth,valign=c]{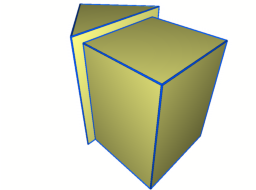} & \includegraphics[width=0.091\linewidth,valign=c]{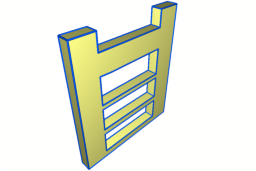} \\
    GT & \includegraphics[width=0.091\linewidth,valign=c]{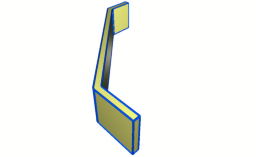} & \includegraphics[width=0.091\linewidth,valign=c]{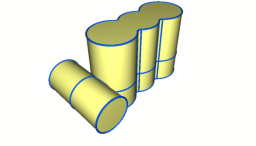} & \includegraphics[width=0.091\linewidth,valign=c]{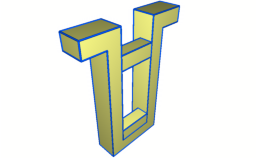} & \includegraphics[width=0.091\linewidth,valign=c]{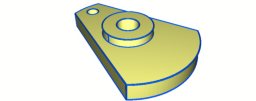} & \includegraphics[width=0.091\linewidth,valign=c]{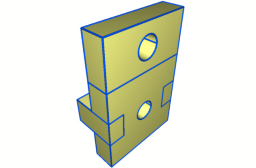} &  \includegraphics[width=0.091\linewidth,valign=c]{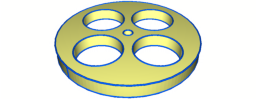} & \includegraphics[width=0.091\linewidth,valign=c]{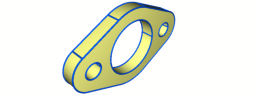} & \includegraphics[width=0.091\linewidth,valign=c]{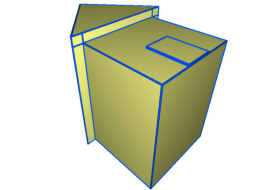} &
    \includegraphics[width=0.091\linewidth,valign=c]{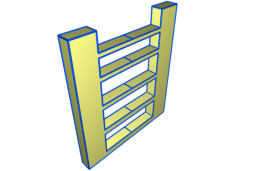} \\
    \end{tabular}
    \caption{Qualitative results on the DeepCAD dataset.}
    \label{fig:example_deepcad}
\end{figure*}

\begin{figure*}[!htb]
    \setlength{\tabcolsep}{0pt}
    \centering
    \begin{tabular}{rccccccccc}
    Point Cloud & \includegraphics[width=0.091\linewidth,valign=c]{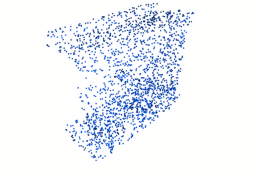} &  \includegraphics[width=0.091\linewidth,valign=c]{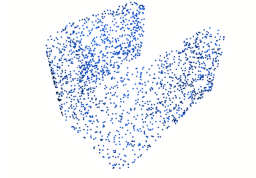} & \includegraphics[width=0.091\linewidth,valign=c]{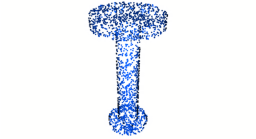} & \includegraphics[width=0.091\linewidth,valign=c]{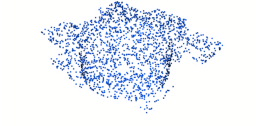} & \includegraphics[width=0.091\linewidth,valign=c]{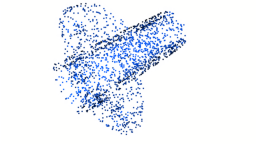} & \includegraphics[width=0.091\linewidth,valign=c]{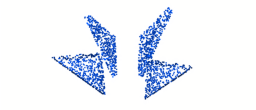} & \includegraphics[width=0.091\linewidth,valign=c]{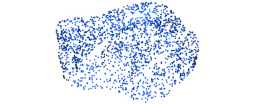} & \includegraphics[width=0.091\linewidth,valign=c]{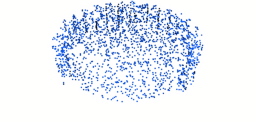} & \includegraphics[width=0.091\linewidth,valign=c]{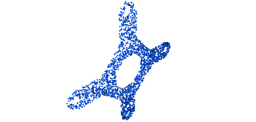} \\
    CAD-SIGNet & \includegraphics[width=0.091\linewidth,valign=c]{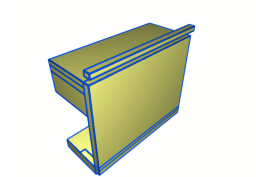} &  \includegraphics[width=0.091\linewidth,valign=c]{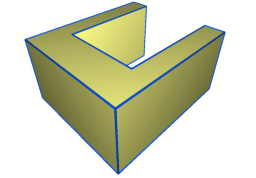} & \includegraphics[width=0.091\linewidth,valign=c]{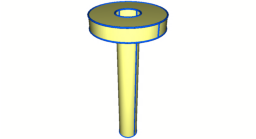} & \includegraphics[width=0.091\linewidth,valign=c]{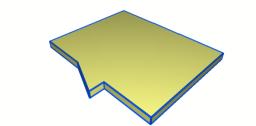} & \includegraphics[width=0.091\linewidth,valign=c]{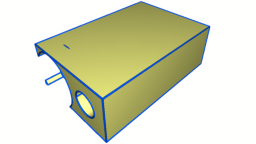} & \includegraphics[width=0.091\linewidth,valign=c]{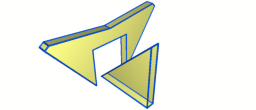} & \includegraphics[width=0.091\linewidth,valign=c]{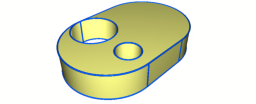} & \includegraphics[width=0.091\linewidth,valign=c]{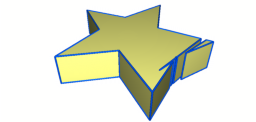} & \includegraphics[width=0.091\linewidth,valign=c]{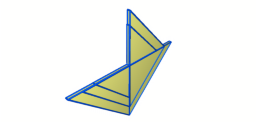} \\
    \modelname & \includegraphics[width=0.091\linewidth,valign=c]{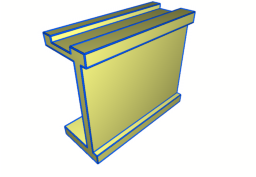} &  \includegraphics[width=0.091\linewidth,valign=c]{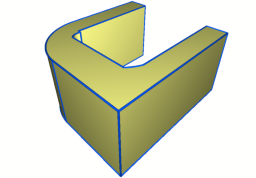} & \includegraphics[width=0.091\linewidth,valign=c]{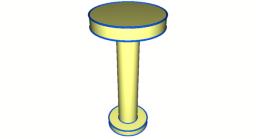} & \includegraphics[width=0.091\linewidth,valign=c]{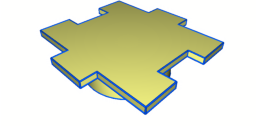} & \includegraphics[width=0.091\linewidth,valign=c]{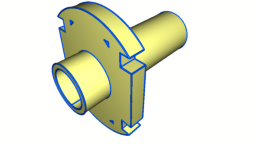} & \includegraphics[width=0.091\linewidth,valign=c]{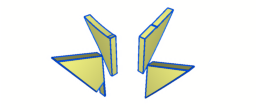} & \includegraphics[width=0.091\linewidth,valign=c]{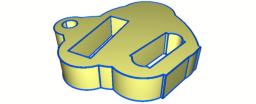} & \includegraphics[width=0.091\linewidth,valign=c]{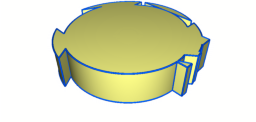} & \includegraphics[width=0.091\linewidth,valign=c]{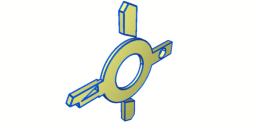} \\
    GT & \includegraphics[width=0.091\linewidth,valign=c]{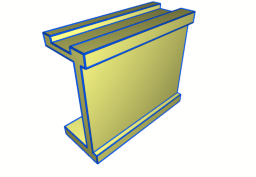} &  \includegraphics[width=0.091\linewidth,valign=c]{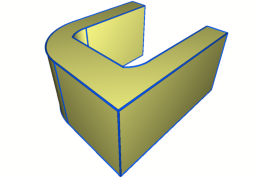} & \includegraphics[width=0.091\linewidth,valign=c]{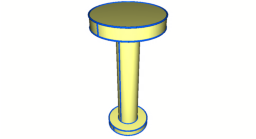} & \includegraphics[width=0.091\linewidth,valign=c]{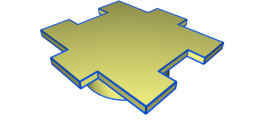} & \includegraphics[width=0.091\linewidth,valign=c]{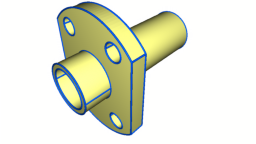} & \includegraphics[width=0.091\linewidth,valign=c]{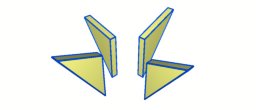} & \includegraphics[width=0.091\linewidth,valign=c]{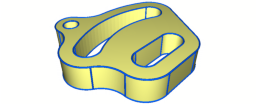} & \includegraphics[width=0.091\linewidth,valign=c]{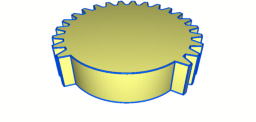} & \includegraphics[width=0.091\linewidth,valign=c]{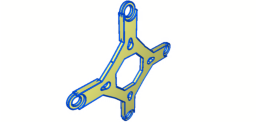} \\
    \end{tabular}
    \caption{Qualitative results on the Fusion360 dataset.}
    \label{fig:example_fusion360}
\end{figure*}

\begin{figure*}[!htb]
    \setlength{\tabcolsep}{0pt}
    \centering
    \begin{tabular}{rccccccccc}
    Point Cloud & \includegraphics[width=0.091\linewidth,valign=c]{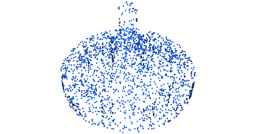} & \includegraphics[width=0.091\linewidth,valign=c]{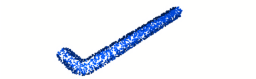} & \includegraphics[width=0.091\linewidth,valign=c]{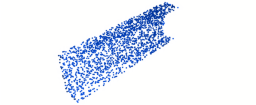} & \includegraphics[width=0.091\linewidth,valign=c]{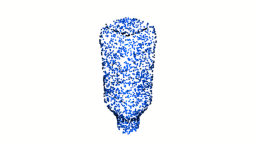} & \includegraphics[width=0.091\linewidth,valign=c]{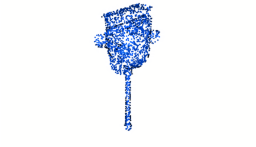} & \includegraphics[width=0.091\linewidth,valign=c]{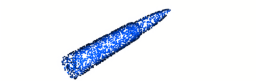} &  \includegraphics[width=0.091\linewidth,valign=c]{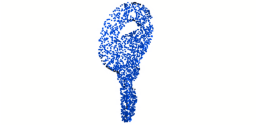} & \includegraphics[width=0.091\linewidth,valign=c]{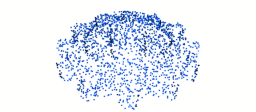} & \includegraphics[width=0.091\linewidth,valign=c]{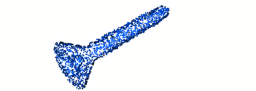} \\
    CAD-SIGNet & \includegraphics[width=0.091\linewidth,valign=c]{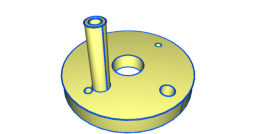} & \includegraphics[width=0.091\linewidth,valign=c]{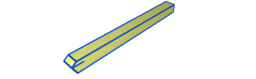} & \includegraphics[width=0.091\linewidth,valign=c]{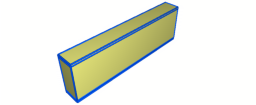} & \includegraphics[width=0.091\linewidth,valign=c]{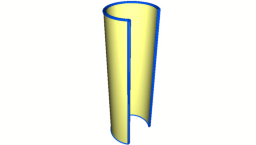} & \includegraphics[width=0.091\linewidth,valign=c]{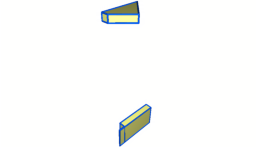} & \includegraphics[width=0.091\linewidth,valign=c]{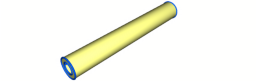} &  \includegraphics[width=0.091\linewidth,valign=c]{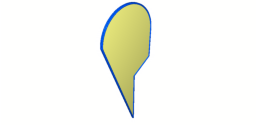} & \includegraphics[width=0.091\linewidth,valign=c]{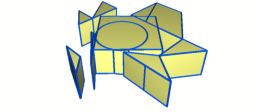} & \includegraphics[width=0.091\linewidth,valign=c]{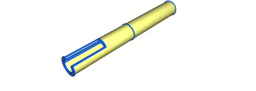} \\
    \modelname & \includegraphics[width=0.091\linewidth,valign=c]{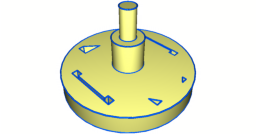} & \includegraphics[width=0.091\linewidth,valign=c]{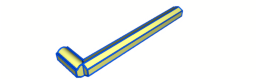} & \includegraphics[width=0.091\linewidth,valign=c]{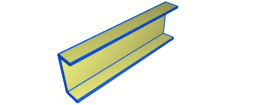} & \includegraphics[width=0.091\linewidth,valign=c]{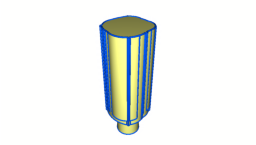} & \includegraphics[width=0.091\linewidth,valign=c]{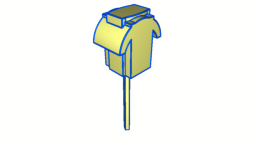} & \includegraphics[width=0.091\linewidth,valign=c]{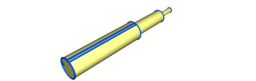} &  \includegraphics[width=0.091\linewidth,valign=c]{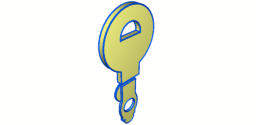} &  \includegraphics[width=0.091\linewidth,valign=c]{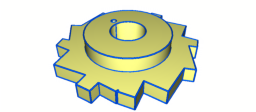} & \includegraphics[width=0.091\linewidth,valign=c]{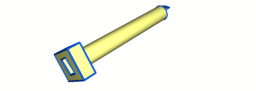} \\
    GT & \includegraphics[width=0.091\linewidth,valign=c]{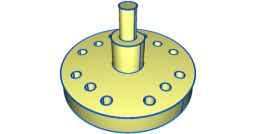} & \includegraphics[width=0.091\linewidth,valign=c]{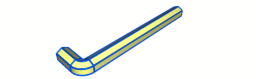} & \includegraphics[width=0.091\linewidth,valign=c]{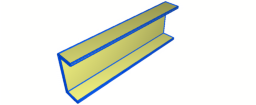} & \includegraphics[width=0.091\linewidth,valign=c]{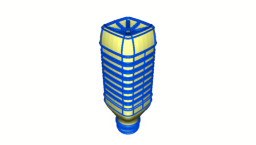} & \includegraphics[width=0.091\linewidth,valign=c]{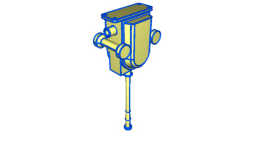} & \includegraphics[width=0.091\linewidth,valign=c]{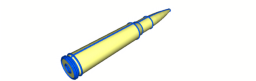} &  \includegraphics[width=0.091\linewidth,valign=c]{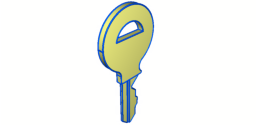} & \includegraphics[width=0.091\linewidth,valign=c]{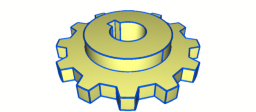} & \includegraphics[width=0.091\linewidth,valign=c]{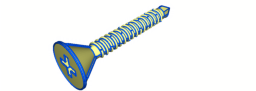} \\
    \end{tabular}
    \caption{Qualitative results on the real-world CC3D dataset.}
    \label{fig:example_cc3d}
\end{figure*}

\noindent \textbf{Qualitative Results:} Additional qualitative results for the reverse engineering of CAD models from point clouds are presented for DeepCAD (\Cref{fig:example_deepcad}), Fusion360 (\Cref{fig:example_fusion360}), and real-world CC3D (\Cref{fig:example_cc3d}) datasets. As detailed in Section~\textcolor{cvprblue}{5.1} of the main paper, \modelname~consistently generates shapes that closely approximate the input point cloud geometry, whereas CAD-SIGNet~\cite{khan2024cad} can generate predictions that greatly differ from the input. 

\begin{figure*}
\begin{tabular}{cc}
\includegraphics[width=0.091\linewidth,valign=c]{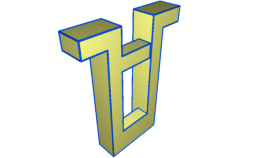} &
\begin{minipage}[c]{0.85\linewidth}
\begin{pythoncode}
import cadquery as cq
w0 = cq.Workplane('XY', origin=(0, 0, -17))
r = w0.sketch().segment((-48, -64), (24, -64)).segment((24, -43)).segment((-27, -43))
    .segment((-27, 60)).segment((27, 60)).segment((27, -39)).segment((48, -39))
    .segment((48, 100)).segment((-48, 100)).close().assemble().finalize().extrude(20)
  .union(w0.sketch().segment((-82, -100),(-27, -100)).segment((-27, 80)).segment((27, 80))
    .segment((27, -100)).segment((82, -100)).segment((82, -79)).segment((48, -79))
    .segment((48, 100)).segment((-48, 100)).segment((-48, -79)).segment((-82, -79))
    .close().assemble().finalize().extrude(34))
\end{pythoncode}
\end{minipage} \\ \\
\includegraphics[width=0.091\linewidth,valign=c]{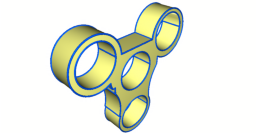} &
\begin{minipage}[c]{0.85\linewidth}
\begin{pythoncode}
import cadquery as cq
w0 = cq.Workplane('XY', origin=(0, 0, -16))
r = w0.sketch().arc((-46, -23), (-95, -74), (-27, -56)).segment((30, -56))
    .arc((96, -72), (44, -25)).segment((44, -12)).arc((31, 14), (30, 42))
    .arc((1, 92), (-31, 44)).arc((-32, 43), (-33, 43)).arc((-31, 20),(-39, -2))
    .segment((-39, -12)).segment((-43, -12)).arc((-45, -17),(-46, -23)).assemble()
    .push([(-64, -56)]).circle(28, mode='s').push([(0, 56)]).circle(28, mode='s')
    .push([(0, -19)]).circle(28, mode='s').push([(65, -56)])
    .circle(28, mode='s').finalize().extrude(32)
\end{pythoncode}
\end{minipage} \\ \\
\includegraphics[width=0.091\linewidth,valign=c]{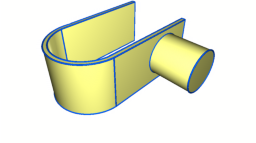} &
\begin{minipage}[c]{0.85\linewidth}
\begin{pythoncode}
import cadquery as cq
w0 = cq.Workplane('ZX', origin=(0, 40, 0))
w1 = cq.Workplane('XY', origin=(0, 0, -19))
r = w0.sketch().arc((-24, -47), (41, -99),(87, -32)).segment((88, -32)).segment((88, 100))
    .segment((82, 100)).segment((82, -52)).arc((34, -94), (-18, -52)).segment((-18, 100))
    .segment((-24, 100)).close().assemble().finalize().extrude(-80)
  .union(w1.workplane(offset=-69 / 2).moveTo(52, 0).cylinder(69, 32))
\end{pythoncode}
\end{minipage} \\ \\
\includegraphics[width=0.091\linewidth,valign=c]{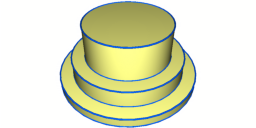} &
\begin{minipage}[c]{0.85\linewidth}
\begin{pythoncode}
import cadquery as cq
w0 = cq.Workplane('ZX', origin=(0, 31, 0))
r = w0.workplane(offset=-75 / 2).cylinder(75, 62)
  .union(w0.workplane(offset=-25 / 2).cylinder(25, 81))
  .union(w0.workplane(offset=13 / 2).cylinder(13, 100))
\end{pythoncode}
\end{minipage} \\ \\
\includegraphics[width=0.091\linewidth,valign=c]{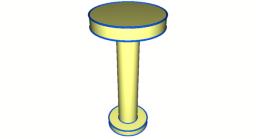} &
\begin{minipage}[c]{0.85\linewidth}
\begin{pythoncode}
import cadquery as cq
w0 = cq.Workplane('ZX', origin=(0, 69, 0))
w1 = cq.Workplane('ZX', origin=(0, -85, 0))
r = w0.workplane(offset=-150 / 2).cylinder(150, 15)
  .union(w0.workplane(offset=10 / 2).cylinder(10, 31))
  .union(w0.workplane(offset=31 / 2).cylinder(31, 8))
  .union(w1.workplane(offset=-15/2).cylinder(15,46))
\end{pythoncode}
\end{minipage} \\ \\
\includegraphics[width=0.091\linewidth,valign=c]{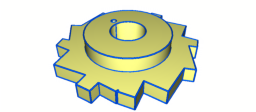} &
\begin{minipage}[c]{0.85\linewidth}
\begin{pythoncode}
import cadquery as cq
w0 = cq.Workplane('ZX', origin=(0, 20, 0))
r = w0.sketch().circle(61).circle(25, mode='s').push([(34, 4)])
    .circle(4, mode='s').finalize().extrude(-41)
  .union(w0.sketch().segment((-100, 19), (-88, 11)).segment((-97, -34)).segment((-67, -41))
    .segment((-77, -66)).segment((-57, -74)).segment((-57, -72)).segment((-56, -72))
    .segment((-56, -75)).segment((-32, -80)).segment((-35, -95)).segment((-16, -100))
    .segment((-11, -83)).segment((33, -100)).segment((45, -70)).segment((68, -76))
    .segment((76, -61)).segment((66, -56)).segment((100, -30)).segment((88, -19))
    .segment((97, 34)).segment((67, 41)).segment((77, 66)).segment((57, 74))
    .segment((51, 69)).segment((51, 70)).segment((32, 76)).segment((35, 95))
    .segment((16, 100)).segment((11, 83)).segment((-33, 100)).segment((-45, 70))
    .segment((-68, 77)).segment((-76, 62)).segment((-66, 56)).close().assemble()
    .circle(26, mode='s').finalize().extrude(-20))
\end{pythoncode}
\end{minipage}
\end{tabular}
\caption{\modelname~predictions on DeepCAD (top $2$ rows), Fusion360 (mid $3$ rows), and CC3D (last row) datasets. Each row contains predicted CadQuery Python code and its result after execution in Python interpreter.}
\label{fig:predictions_code}
\end{figure*}

\noindent \textbf{Code Outputs:} \Cref{fig:predictions_code} illustrates the predicted code sequences and their corresponding reconstructed shapes. The predicted codes have a syntax that is consistent with the procedurally generated training examples, showing that \modelname~successfully learns both the features and CAD design patterns established in the training set.

\noindent \textbf{Ablation Results:} \Cref{tab:ablations_supp} shows the architecture ablation results on all metrics, complementing Table~\textcolor{iccvblue}{4} of the main paper. Results show that for the same size of input point clouds Qwen1.5b always produces better geometric performance (median CD and IoU) than Qwen0.5b. This can be attributed to the higher number of parameters as well as to the better ability of the model to produce valid python code before fine-tuning. Furthermore, increasing the size of the input point cloud demonstrates a similar pattern, with Qwen1.5b with an $256$ input points appears to be the set of architecture parameters leading to the best performance. Note that the mean CD is a metric that is very sensitive to outlier predictions. While Qwen1.5b with $256$ input points appears to result in the highest IR, it is negligibly low on all datasets (less $0.5\%$). This can also be explained by the fact that this setting produces more complex CAD sketch-extrude sequences, making them more susceptible to errors. Note that a key idea of our method is to leverage pre-trained LLMs as decoder of Python code. In the absence of LLM-based CAD reverse engineering methods, we compare our approach to SOTA methods despite the difference in model sizes. For reference, CAD-SIGNet contains $6$\, M parameters.  

\begin{table*}
\centering
\resizebox{\linewidth}{!}{
\begin{tabular}{cccccccccccccc}
\toprule
\multirow{2}{*}{Points} & Model & \multicolumn{4}{c}{DeepCAD} & \multicolumn{4}{c}{Fusion360} & \multicolumn{4}{c}{CC3D} \\
& Size & Mean CD\textdownarrow & Med. CD\textdownarrow & IoU\textuparrow & IR\textdownarrow & Mean CD\textdownarrow & Med. CD\textdownarrow & IoU\textuparrow & IR\textdownarrow & Mean CD\textdownarrow & Med. CD\textdownarrow & IoU\textuparrow & IR\textdownarrow \\
\midrule
\multirow{2}{*}{$64$} & $0.5$\,B & $0.42$ & $0.20$ & $88.5$ & $0.1$ & $0.58$ & $0.22$ & $82.1$ & $0.1$ & $0.87$ & $0.45$ & $70.1$ & $0.1$ \\
& $1.5$\,B & $0.36$ & $0.19$ & $89.3$ & $0.0$ & $0.43$ & $0.20$ & $83.7$ & $0.1$ & $0.83$ & $0.42$ & $71.2$ & $0.0$ \\
\midrule
\multirow{2}{*}{$128$} & $0.5$\,B & $0.36$ & $0.18$ & $89.9$ & $0.1$ & $0.43$ & $0.18$ & $84.3$ & $0.1$ & $0.87$ & $0.38$ & $71.9$ & $0.1$ \\
& $1.5$\,B & $\mathbf{0.27}$ & $0.17$ & $91.0$ & $0.1$ & $0.36$ & $0.17$ & $86.1$ & $0.1$ & $0.79$ & $0.34$ & $73.1$ & $0.1$ \\
\midrule
\multirow{2}{*}{$256$} & $0.5$\,B & $0.36$ & $0.17$ & $90.6$ & $0.2$ & $0.40$ & $0.17$ & $85.4$ & $0.4$ & $0.87$ & $0.36$ & $72.6$ & $0.1$ \\
& $1.5$\,B & $0.30$ & $\mathbf{0.16}$ & $\mathbf{92.0}$ & $0.4$ & $\mathbf{0.35}$ & $\mathbf{0.15}$ & $\mathbf{87.8}$ & $0.5$ & $\mathbf{0.76}$ & $\mathbf{0.31}$ & $\mathbf{74.2}$ & 0.3\\
\bottomrule
\end{tabular}}
\caption{Ablation of architecture details.}
\label{tab:ablations_supp}
\end{table*}

\noindent \textbf{Command \& Parameter Accuracy:} In order to evaluate the ability of \modelname~to predict numerical values and sequences that are consistent with the training set, we evaluate \modelname~trained on the DeepCAD dataset converted to CADQuery python codes with the Acc\textsubscript{command} and Acc\textsubscript{parameter} as introduced in~\cite{wu2021deepcad}. The results on the DeepCAD testing set are presented in~\Cref{tab:acc}. It can be observed that \modelname~achieves comparable performance to the state-of-the-art on the commnad type accuracy and significantly higher performance on the parameter accuracy. This demonstrates that \modelname~is able to predict numerical values accurately. Note that, those metrics were originally developed to evaluate autoencoding ability. However, there may exist many different possible valid CAD sequences to construct the same CAD model and these metrics do not take this into account. As a result, these metrics were omitted in recent works (CAD-SIGNet~\cite{khan2024cad} and TransCAD~\cite{dupont2024transcad}).

\vspace{1em}

\begin{figure*}
\begin{tabular}{cc}
\includegraphics[width=0.091\linewidth,valign=c]{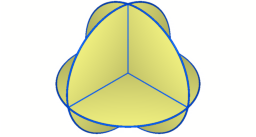} &
\begin{minipage}[c]{0.85\linewidth}
\begin{pythoncode}
import cadquery as cq
w0 = cq.Workplane('XY', origin=(0, 0, 0))
w1 = cq.Workplane('YZ', origin=(0, 0, 0))
r = w0.workplane(offset=0 / 2).cylinder(@\highlight{turquoise!50}{0}@, 98)
  .union(w1.workplane(offset=0 / 2).cylinder(@\highlight{turquoise!50}{0}@, 100))
\end{pythoncode}
\end{minipage}
\end{tabular}
\small (a) The ground truth model contains three very thin cylinders with height smaller than $1$. As a result, \modelname~is not able to predict heights with sufficient precision due to quantization and predicts cylinders with height $0$, producing an invalid model. 
\\ \\
\begin{tabular}{cc}
\includegraphics[width=0.091\linewidth,valign=c]{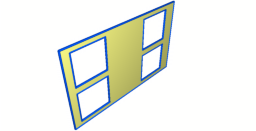} &
\begin{minipage}[c]{0.85\linewidth}
\begin{pythoncode}
import cadquery as cq
w0 = cq.Workplane('XY', origin=(0, 0, 0))
r = w0.sketch().rect(200, 124).push([(-63.5, 25)]).rect(51, 60, mode='s')
    .push([(55, -25)]).rect(50, 60, mode='s').finalize()@\highlight{turquoise!50}{.extrude(0)}@
\end{pythoncode}
\end{minipage}
\end{tabular}
\small (b) As the ground-truth model has thickness less than $1$, \modelname~predicts an extrusion distance of $0$ as a quantized approximation (highlighted in yellow), resulting in an invalid CAD model. \\ \\
\begin{tabular}{cc}
\includegraphics[width=0.091\linewidth,valign=c]{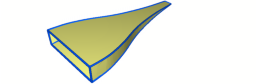} &
\begin{minipage}[c]{0.85\linewidth}
\begin{pythoncode}
import cadquery as cq
w0 = cq.Workplane('YZ', origin=(34, 0, 0))
w1 = cq.Workplane('XY', origin=(0, 0, 44))
r = w0.sketch().segment((-7, -35), (11, -36)).segment((11, -24)).arc((1, -14),
    (6, -2)).segment((-1, 19)).segment((11, 23)).segment((11, 28))
    .segment((11, 29)).segment((12, 29)).segment((12, 35))
    .segment((-4, 36)).close().assemble().finalize().extrude(-133)
  .union(w0.sketch().segment((5, -7), (14, -2)).segment((8, 8)).arc((7, 0),
    (5, -7)).assemble().finalize().extrude(63))
  .union(w1.sketch().arc((-100, 12), (-85, 10), (-70, 5)).arc((-68, 6),
    (-66, 5)).arc((-59, 4), (-52, 2)).@\highlight{turquoise!50}{arc((-51, 3), (-50, 4))}@.arc((-72, 7),
    (-90, 12)).close().assemble().finalize().extrude(-88))
\end{pythoncode}
\end{minipage}
\end{tabular}
\small (c) The ground-truth CAD model is created with B-spline primitives. Since \modelname~supports only arc, circle and line primitives, it tries to approximate the solution with multiple arcs, but fails to provide a valid CAD model. In particular, the prediction contains an arc constructed from three co-linear points (highlighted in yellow), which raises an error in CadQuery.\\ \\
\begin{tabular}{cc}
\includegraphics[width=0.091\linewidth,valign=c]{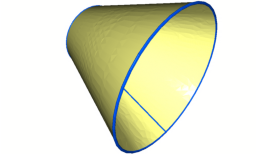} &
\begin{minipage}[c]{0.85\linewidth}
\begin{pythoncode}
import cadquery as cq
w0 = cq.Workplane('XY', origin=(0, 0, -79))
r = w0.sketch().segment((-100, -1), (-91, -1)).arc((0, -93),(91, -1))
    .segment((100, -1)).segment((100, 1)).segment((91, 1)).arc((0, 99),
    (-91, 1)).segment((-100, 1)).close().assemble().push([(0, -2)])
    .circle(90, mode='s').finalize().extrude(-2)
  .union(w0.workplane(offset=140 / 2).cylinder(140, 72))
  .union(@\highlight{turquoise!50}{w0.sketch().segment((-51, 15), (-50, 15)).arc((0, -53),(50, 15))
    .segment((51, 15)).segment((51, 27)).segment((48, 27)).arc((0, -53),
    (-48, 27)).segment((-51, 27))}@.close().assemble().finalize().extrude(159))
\end{pythoncode}
\end{minipage}
\end{tabular}
\small (d)  The ground-truth CAD model is created with a revolution operation. Since \modelname~supports only extrusion operation, it tries to approximate the solution with multiple arcs. However, one of the sketch (highlighted in yellow) results in a self-intersecting loop, which is not a valid face.
\caption{Examples of invalid predictions. Each row contains the ground-truth CAD model (left) and an invalid predicted CadQuery Python code (right). The CAD models in (a) and (b) are taken from the DeepCAD dataset and the CC3D dataset for (c) and (d). Invalid predictions mostly take place when the ground-truth contains features of very small dimension with respect to the size of the CAD model as in (a) and (b), or when the ground-truth model contains operations other than the ones supported as in (c) and (d).}
\label{fig:failure}
\end{figure*}

\noindent \textbf{Invalid Predictions:} The invalidity rate of \modelname~predictions is very low, below $1\%$ on the DeepCAD~\cite{wu2021deepcad}, Fusion360~\cite{willis2021fusion} and real-world CC3D~\cite{mallis2023sharp} dataset. Some examples of invalid code predictions are presented in \Cref{fig:failure}. Invalid predictions happen when the CAD model contains features of dimension smaller than the resolution induced by quantization (\Cref{fig:failure}\textcolor{cvprblue}{(a)} and \textcolor{cvprblue}{(b)}) or when the ground truth CAD model contains features, such as revolution or B-spline, that are not present in the training dataset (\Cref{fig:failure}\textcolor{cvprblue}{(c)} and \textcolor{cvprblue}{(d)})).

\begin{table}[th!]
\centering
\resizebox{0.85\linewidth}{!}{\begin{tabular}{lcc}
\toprule
Method  & Acc\textsubscript{command} (\%) & Acc\textsubscript{parameter} (\%) \\
\midrule
DeepCAD~\cite{wu2021deepcad}  & 80.4 & 69.6 \\
PrismCAD~\cite{lambourne2022reconstructing}  & 73.0 & 66.8 \\
HNC-CAD~\cite{xu2023hnc}  & 82.7 & 74.6 \\
CAD-Diffuser~\cite{ma2024draw} & 88.5 & 82.9 \\
\modelname  & 83.9 & 92.1 \\
\bottomrule
\end{tabular}}
\caption{Command and parameter accuracy results~\cite{wu2021deepcad} on the DeepCAD dataset. All methods (incl. \modelname) are trained and tested on DeepCAD dataset.}
\label{tab:acc}
\end{table}

\clearpage

\clearpage
\clearpage

\clearpage

\section{Test-time Sampling}
\begin{figure*}
    \setlength{\tabcolsep}{0pt}
    \centering \footnotesize
    \begin{tabular}{ccccccc}
    Point Cloud & GT & \multicolumn{5}{c}{\modelname~Predictions} \\
    \includegraphics[width=0.091\linewidth,valign=c]{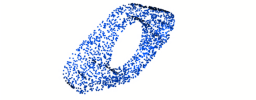} & 
    \includegraphics[width=0.091\linewidth,valign=c]{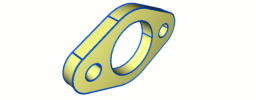} &
    \includegraphics[width=0.091\linewidth,valign=c]{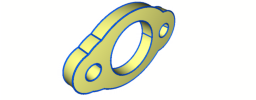} & 
    \includegraphics[width=0.091\linewidth,valign=c]{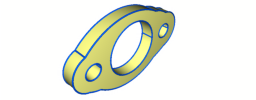} &
    \includegraphics[width=0.091\linewidth,valign=c]{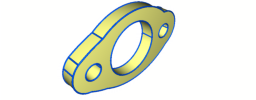} &
    \includegraphics[width=0.091\linewidth,valign=c]{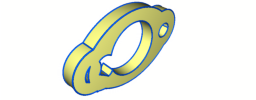} &
    \includegraphics[width=0.091\linewidth,valign=c]{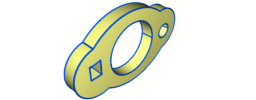} \\
    \includegraphics[width=0.091\linewidth,valign=c]{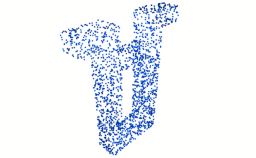} & \includegraphics[width=0.091\linewidth,valign=c]{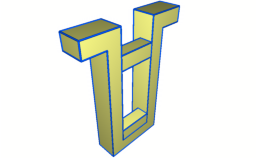} & \includegraphics[width=0.091\linewidth,valign=c]{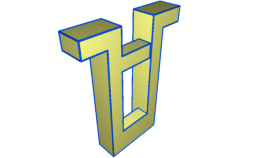} & \includegraphics[width=0.091\linewidth,valign=c]{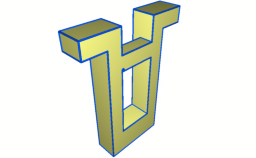} & \includegraphics[width=0.091\linewidth,valign=c]{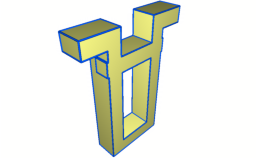} & \includegraphics[width=0.091\linewidth,valign=c]{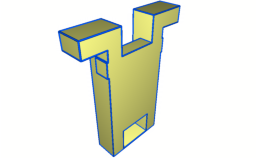} & \includegraphics[width=0.091\linewidth,valign=c]{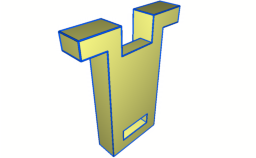} \\
    \includegraphics[width=0.091\linewidth,valign=c]{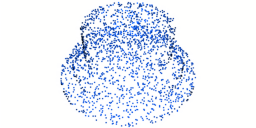} &
    \includegraphics[width=0.091\linewidth,valign=c]{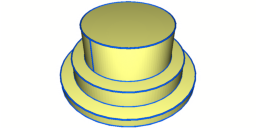} &
    \includegraphics[width=0.091\linewidth,valign=c]{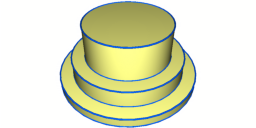} &
    \includegraphics[width=0.091\linewidth,valign=c]{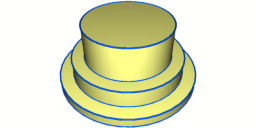} &
    \includegraphics[width=0.091\linewidth,valign=c]{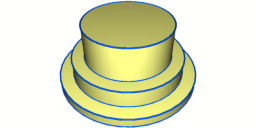} &
    \includegraphics[width=0.091\linewidth,valign=c]{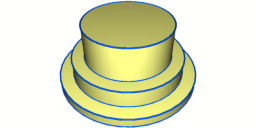} &
    \includegraphics[width=0.091\linewidth,valign=c]{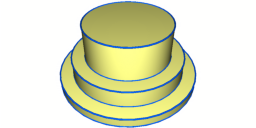} \\
    \includegraphics[width=0.091\linewidth,valign=c]{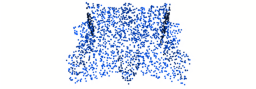} &
    \includegraphics[width=0.091\linewidth,valign=c]{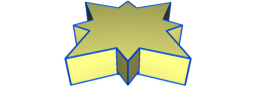} &
    \includegraphics[width=0.091\linewidth,valign=c]{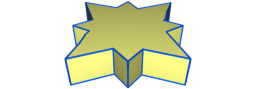} &
    \includegraphics[width=0.091\linewidth,valign=c]{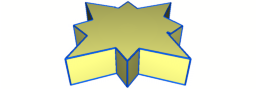} &
    \includegraphics[width=0.091\linewidth,valign=c]{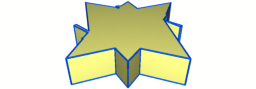} &
    \includegraphics[width=0.091\linewidth,valign=c]{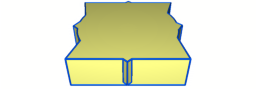} &
    \includegraphics[width=0.091\linewidth,valign=c]{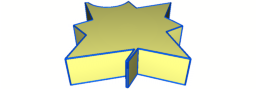} \\
    \includegraphics[width=0.091\linewidth,valign=c]{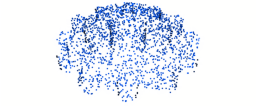} & \includegraphics[width=0.091\linewidth,valign=c]{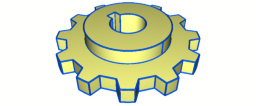} & \includegraphics[width=0.091\linewidth,valign=c]{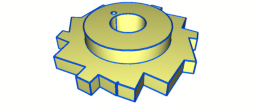} & \includegraphics[width=0.091\linewidth,valign=c]{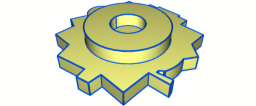} & \includegraphics[width=0.091\linewidth,valign=c]{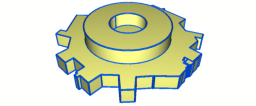} & \includegraphics[width=0.091\linewidth,valign=c]{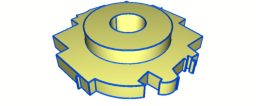} & \includegraphics[width=0.091\linewidth,valign=c]{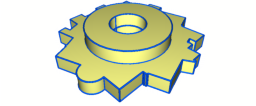} \\
    \includegraphics[width=0.091\linewidth,valign=c]{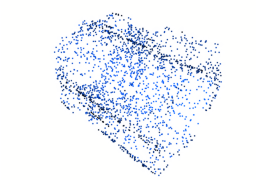} & \includegraphics[width=0.091\linewidth,valign=c]{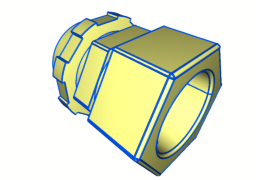} & \includegraphics[width=0.091\linewidth,valign=c]{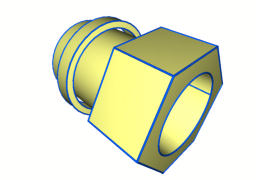} & \includegraphics[width=0.091\linewidth,valign=c]{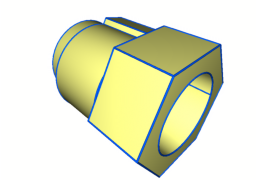} & \includegraphics[width=0.091\linewidth,valign=c]{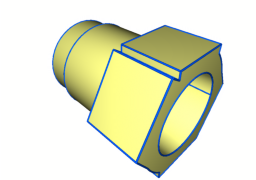} & \includegraphics[width=0.091\linewidth,valign=c]{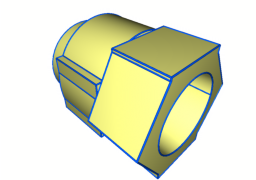} & \includegraphics[width=0.091\linewidth,valign=c]{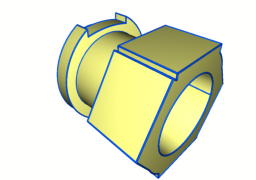}
    \end{tabular}
    \caption{\modelname~predictions from different point cloud sampling on DeepCAD, Fusion360, and real-world CC3D datasets. For each prediction, $256$ points are sampled randomly from the input point cloud.}
    \label{fig:sampling}
\end{figure*}

The ablation study in Section~\textcolor{cvprblue}{5.1} of the main paper demonstrates the effectiveness of our test-time sampling strategy. This approach generates multiple plausible solutions by sampling different input point clouds. \Cref{fig:sampling} illustrates the qualitative results from different sampling instances. While \modelname~successfully captures the overall geometry across different samplings, fine-grained details may vary in reconstruction quality due to the relatively sparse point cloud input. However, this limitation can be effectively addressed by leveraging multiple sampling iterations to capture different aspects of the input geometry.

\section{Interpretability and CAD-QA}

In this section, we provide further details on the CAD-QA experiments reported in Section~\textcolor{cvprblue}{5.2} of the main paper. We start by providing more details on the SGP-Bench benchmark~\cite{qiu2024can}. Then, we present results further results and examples of GPT-4o outputs.

\subsection{Representation and CAD-QA}

\begin{figure*}
    \setlength{\tabcolsep}{0pt}
    \centering
    \begin{tabular}{lc} %
    \raisebox{-\height}{\includegraphics[width=0.091\linewidth]{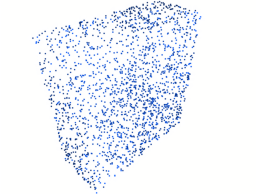}}
    & \fbox{\parbox[t]{0.85\linewidth}{%
    \footnotesize
Examine the following CAD code carefully to understand the 3D object it generates and answer the question based on your interpretation of the rendered image of that object.
\\   \\     
SOL; Line:(221,128); Line:(221,223) ;Line:(128,223); Line:(128,128); Ext: (128,128,128,32,110,128,98,167,128, Newbody, One-sided); EOS
\\ \\
\textbf{Hint:} the CAD code has the following syntax:
CAD code consists of a sequence of CAD commands that describe a 3D object.
The commands fall into two categories: sketch and extrusion. 
Sketch commands are used to specify closed curves on a 2D plane in 3D space. Each closed curve is referred as a loop, and one or more loops form a closed region called a profile. A loop always starts with an indicator command {\textless}SOL{\textgreater} followed by a series of curve commands. All the curves on the loop are in counterclockwise order, beginning with the curve whose starting point is at the most bottom-left. In total, there are three possible curve commands: Line, Arc, and Circle. Line(x, y): a line, with x, y as line end-point. Arc(x, y, u, f): an arc, with x,y as arc end-point, u as sweep angle and f as whether it is counter-clockwise, f=0 means it is counter-clockwise, f=1 means it is not counter-clockwise. Circle(x, y, r): a circle, with x,y as the center point and r as the radius. The extrusion command has two purposes: 1) It extrudes a sketch profile from a 2D plane into a 3D body, and the extrusion type can be either one-sided, symmetric, or two-sided with respect to the profile's sketch plane. 2) The command also specifies (through the parameter b in Ext) how to merge the newly extruded 3D body with the previously created shape by one of the boolean operations: either creating a new body, or joining, cutting or intersecting with the existing body. Ext(x, y, z, o, p, q, s, e, f, b, u): extrude operation, with x, y, z as the sketch plane orientation, o, p, q as the sketch plane origin, s as the scale of the associated sketch profile, e, f as the extrude distances towards both sides, b as the type of merge operation (could be New-body operation, join operation, cut operation and intersect operation) and u as the extrude type (could be one-sided, symmetric or two-sided). {\textless}EOS{\textgreater} means the end of the code. 
\\

\textbf{Question:} How many faces does the CAD object in the image have?
    }} 
    \\
    & \small (a) DeepCAD Representation
    \\ \addlinespace[1em]
     
    & \fbox{\parbox{0.85\linewidth}{%
    \footnotesize
        Examine the following CAD code carefully to understand the 3D object it generates and answer the question based on your interpretation of the rendered image of that object.
\\ \\
import cadquery as cq

def make\_shape():

	\qquad plane0 = cq.Plane(origin = (-0.75,-0.1406,0.0),xDir = (1.0,0.0,0.0),normal = (0.0,0.0,1.0))
	
   \qquad w0 = cq.Workplane(plane0)
	
    \qquad face0 = w0.sketch().face(w0.sketch().segment( (0.0, 0.0), (0.7495, 0.0)).segment((0.7495, 0.0), (0.7495, 0.7656)).segment((0.7495, 0.7656), (0.0, 0.7656)).segment((0.0, 0.7656), (0.0, 0.0)).assemble(), mode = 'a').finalize()
	
    \qquad shape0 = face0.extrude(0.3046875, both=False)
	
    \qquad return shape0
    \\ \\
\textbf{Question:} How many faces does the CAD object in the image have?
    }} \\
    & \small (b) CadQuery  Representation \\
    \end{tabular}
    \caption{Example of a CAD question from SGP-Bench in the DeepCAD representation (b) and using the CadQuery representation (c). When asking the question in CadQuery format no hint is provided, while in DeepCAD format a long interpretative hint is required. The choice of answers are A) 6, B) 4, C) 8 and D) 5. The correct answer is A) 6.}
    \label{fig:sgp}
\end{figure*}

The goal of the SGP-Bench benchmark is to evaluate the spatial-semantic reasoning skills of LLMs from symbolic graphics programs~\cite{qiu2024can}. One aspect of the benchmark is a set of $1000$ multiple choice questions on 3D CAD models given their corresponding sketch-extrude sequence in the DeepCAD~\cite{wu2021deepcad} format. An example is depicted in \Cref{fig:sgp}. 

\begin{figure*}
    \setlength{\tabcolsep}{0pt}
    \centering
    \begin{tabular}{lc} %
    \raisebox{-\height}{\includegraphics[width=0.091\linewidth]{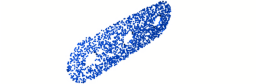}}
    & \fbox{\parbox[t]{0.85\linewidth}{%
\scriptsize
Question: How many holes does the CAD object in the image have?

A) Two, B) Three, C) Four, D) Five

Correct Answer: B
    }} 
    \\
    & \small (a) Point cloud and corresponding CAD-specific question
    \\ \addlinespace[1em]
     
    & \fbox{\parbox{0.85\linewidth}{%
    \scriptsize
The CAD object in the image has five holes.
    }} \\
    & \small (b) PointLLM output from point cloud\\
\addlinespace[1em]
     
    & \fbox{\parbox{0.85\linewidth}{%
    \scriptsize

To determine the number of holes in the CAD object, let's break down the CAD code step by step:

1. \textbf{Extrusion Command:}
   
   Start;Ext:(128,130,128,128,130,128,128,128,NewBody,96);

   This command sets up an extrusion with specific parameters and indicates that a new body will be created.

2. \textbf{First Loop:}
   
   Line:(43,0);Arc:(213,0,255,43);Line:(213,85);Arc:(43,85,0,43);EOL;
   
   \quad- Line:(43,0) - A line from the starting point to (43,0).
   
   \quad - Arc:(213,0,255,43) - An arc with midpoint (213,0) and endpoint (255,43).
   
   \quad- Line:(213,85) - A line from the previous endpoint to (213,85).
   
   \quad- Arc:(43,85,0,43) - An arc with midpoint (43,85) and endpoint (0,43).
   
   \quad - EOL - End of the loop.

   This forms a closed loop, which is a profile for extrusion.

3. \textbf{Second Loop:}

   Circle:(43,43,43,51);EOL;

   \quad- Circle:(43,43,43,51) - A circle with the most left point at (43,43) and the most top point at (43,51).
   
   \quad- EOL - End of the loop.

   This forms another closed loop, which is another profile for extrusion.

4. \textbf{End of Sketch and Extrusion:}
  
EOF;EOS;End;
 
  \quad - EOF - End of face.
 
   \quad- EOS - End of sketch.
   
   \quad- End - End of the code.

From the above analysis, we can see that the CAD object consists of:

\quad- One main profile formed by the first loop.

\quad- One circular profile formed by the second loop.

The circular profile represents a hole within the main profile. Since there is only one circular profile, the CAD object has one hole.

Answer: A

    }} \\
    & \small (c) GPT-4o output from CAD-SIGNet prediction\\
    \addlinespace[1em]
     
    & \fbox{\parbox{0.85\linewidth}{%
    \scriptsize

Let's break down the CAD code step by step to understand the 3D object it generates and determine the number of holes.

1. \textbf{Workplane Creation:}

       w0 = cq.Workplane('XY', origin=(0,0,1))

   This line creates a workplane on the XY plane with an origin at (0,0,1).

2. \textbf{Sketching the Shape:}

   r = w0.sketch()

   This initializes a sketch on the workplane.

3. \textbf{Defining the Shape:}

   .segment((-66,-34),(50,-34))
   .segment((50,-32))
   .arc((100,0),(50,32))
   .segment((50,34))
   .segment((-66,34))
   .segment((-66,32))
   .arc((-100,0),(-66,-32))
   .close()

   These lines define a closed shape with segments and arcs. The shape appears to be symmetrical along the Y-axis.

4. \textbf{Adding Holes:}

   .assemble()
   .push([(-66,0)]).circle(10, mode='s')
   .push([(0,0)]).circle(10, mode='s')
   .push([(66,0)]).circle(10, mode='s')
   .finalize()

   These lines add three circles (holes) to the shape at positions (-66,0), (0,0), and (66,0).

5. \textbf{Extruding the Shape:}

   .extrude(-4)
   
   This line extrudes the shape by -4 units, creating a 3D object with the defined shape and holes.

From the code, we can see that three circles (holes) are added to the shape. Therefore, the CAD object has three holes.

Answer: B

    }} \\
    & \small (d) GPT-4o output from \modelname~prediction \\
    \end{tabular}
    \caption{Example of Point cloud CAD-QA (a) and answers provided by PointLLM (b), CADSIGNet and GPT-4o (c) and \modelname~and GPT-4o (d). 
    }
    \label{fig:pointcloud_sgp}
\end{figure*}

To evaluate the interpretability of our code-based CAD representation, we translated the $1000$ questions of SGP-Bench from the DeepCAD representation (\Cref{fig:sgp}\textcolor{cvprblue}{(a)}) to the CadQuery code format (\Cref{fig:sgp}\textcolor{cvprblue}{(b)}). Using the same protocol as in SGP-Bench~\cite{qiu2024can}, and GPT-4o~\cite{openai2024gpt4technicalreport}, we found that the accuracy on the multiple choice question in CadQuery format is $82.4\%$. This is about $4\%$ higher than using the DeepCAD format with an interpretative hint.  This suggests the proposed code representation provides a more structured and naturally LLM-interpretable representation of CAD models.

\subsection{Point Cloud and CAD-QA}

In Table~\textcolor{cvprblue}{6} of the main paper, the results for point cloud CAD-QA are presented. \Cref{fig:pointcloud_sgp}\textcolor{cvprblue}{(a)} depicts an example of point cloud and question that was used to obtain these results. In this particular question, the task is to deduce the number of holes present in the CAD model given the point cloud as input. \Cref{fig:pointcloud_sgp}\textcolor{cvprblue}{(b)}, the answer provided by PointLLM is shown and it can be observed that PointLLM is unable to retrieve the correct answer. It is worth noting that PointLLM is a network trained to answer semantic questions about object given its point cloud representation, as result in most cases the network is unable to describe geometric CAD-specific questions. For both CAD-SIGNet and \modelname, the point cloud CAD-QA is done in a two step process. First the sketch-extrude is sequence is predicted from each network, then the sequence along with the question is passed through GPT-4o. Note that for CAD-SIGNet an interpretative hint is provided to provide context on the structure of the sequence. A sample output for CAD-SIGNet and GPT-4o can be found in  \Cref{fig:pointcloud_sgp}\textcolor{cvprblue}{(c)}, and in \Cref{fig:pointcloud_sgp}\textcolor{cvprblue}{(d)} for \modelname~and GPT4-o. As the sequence was incorrectly predicted by CAD-SIGNet the answer to the question is wrong (1 hole), whereas the prediction from \modelname~captured better the geometry of the input point cloud leading to a correct answer. It is worth noting, that despite not being provided any information about CadQuery Python code in the prompt, GPT-4o is able to breakdown the predicted sequence into its primitive components and provide correct and accurate geometric descriptions. This can be explained by the fact that LLMs are exposed to large amounts of code data during training. As a result, the CadQuery Python representation of CAD models is appropriate for 

\section{Editing Pipeline Details}

\begin{figure*}[t]
    \setlength{\textfloatsep}{5pt}
    \begin{subfigure}{\textwidth}
        \begin{minted}[
            frame=single, 
            fontsize=\scriptsize,
            baselinestretch=0.8,
            breaklines, 
            breakanywhere, 
            style=friendly
        ]{python}
import cadquery as cq
w0 = cq.Workplane('XY', origin=(0, 0, 0)) 
r = w0.workplane(offset=-100 / 2).cylinder(100, 100)
  .union(w0.workplane(offset=100 / 2).cylinder(100, 34))
        \end{minted}
        \caption{\modelname~output as CadQuery Python code.}
        \label{fig:editing_code}
    \end{subfigure}
    \begin{subfigure}{\textwidth}
        \begin{tcolorbox}[
            width=\textwidth,
            colback=white,
            colframe=black,
            boxrule=0.4pt,
            arc=0pt,
            top=2pt,
            bottom=2pt,
            left=5pt,
            right=5pt
        ]
\scriptsize Refactor the code, add functions and variables that can make the code easier to edit and more interpretable. Add in the code, sliders to edit the size of each primitive: 
import cadquery as cq w0=cq.Workplane('XY',origin=(0,0,0)) r=w0.workplane(offset=-100/2).cylinder(100,100).union(w0.workplane(offset=100/2).cylinder(100,34))
        \end{tcolorbox}
        \caption{Prompt for GPT-4o}
        \label{fig:editing_prompt}
    \end{subfigure}
    \begin{subfigure}{\textwidth}
        \begin{minted}[
            frame=single, 
            fontsize=\tiny,
            baselinestretch=0.8,
            breaklines,  
            style=friendly
        ]{python}
import cadquery as cq
from ipywidgets import interact, FloatSlider

# Default parameters for the CAD model
default_base_diameter = 100
default_base_height = 100
default_top_diameter = 34
default_top_height = 100

def create_cylinder(workplane, diameter, height, offset):
    """Creates a cylinder on the specified workplane with a given diameter, height, and offset."""
    return workplane.workplane(offset=offset).cylinder(height, diameter / 2)

def build_model(base_diameter=default_base_diameter, base_height=default_base_height,
                top_diameter=default_top_diameter, top_height=default_top_height):
    """Builds the model with adjustable base and top cylinder dimensions."""
    # Initialize the main workplane
    w0 = cq.Workplane("XY", origin=(0, 0, 0))
    
    # Create base and top cylinders
    base_cylinder = create_cylinder(w0, base_diameter, base_height, offset=-base_height / 2)
    top_cylinder = create_cylinder(w0, top_diameter, top_height, offset=base_height / 2)
    
    # Union the cylinders
    model = base_cylinder.union(top_cylinder)
    return model

# Function to display the model interactively with sliders for each cylinder dimension
def display_model_with_sliders():
    """Interactive display function with sliders for base and top cylinder dimensions."""
    
    # Sliders for base and top cylinder dimensions
    base_diameter_slider = FloatSlider(value=default_base_diameter, min=10, max=200, step=5, description="Base Diameter")
    base_height_slider = FloatSlider(value=default_base_height, min=10, max=200, step=5, description="Base Height")
    top_diameter_slider = FloatSlider(value=default_top_diameter, min=10, max=200, step=5, description="Top Diameter")
    top_height_slider = FloatSlider(value=default_top_height, min=10, max=200, step=5, description="Top Height")
    
    # Update function to apply slider changes
    @interact(
        base_diameter=base_diameter_slider,
        base_height=base_height_slider,
        top_diameter=top_diameter_slider,
        top_height=top_height_slider
    )
    def update_model(base_diameter, base_height, top_diameter, top_height):
        # Build and display the model
        model = build_model(base_diameter, base_height, top_diameter, top_height)
        show_object(model)  # Use this in CQ-editor or a Jupyter environment

# Execute the function to show sliders and interactive model
display_model_with_sliders()
        \end{minted}
        \caption{GPT-4o output: executable refactored code with interactive sliders}
        \label{fig:editing_output}
    \end{subfigure}
     \begin{subfigure}[b]{\linewidth}
\centering
\includegraphics[width=0.45\linewidth]{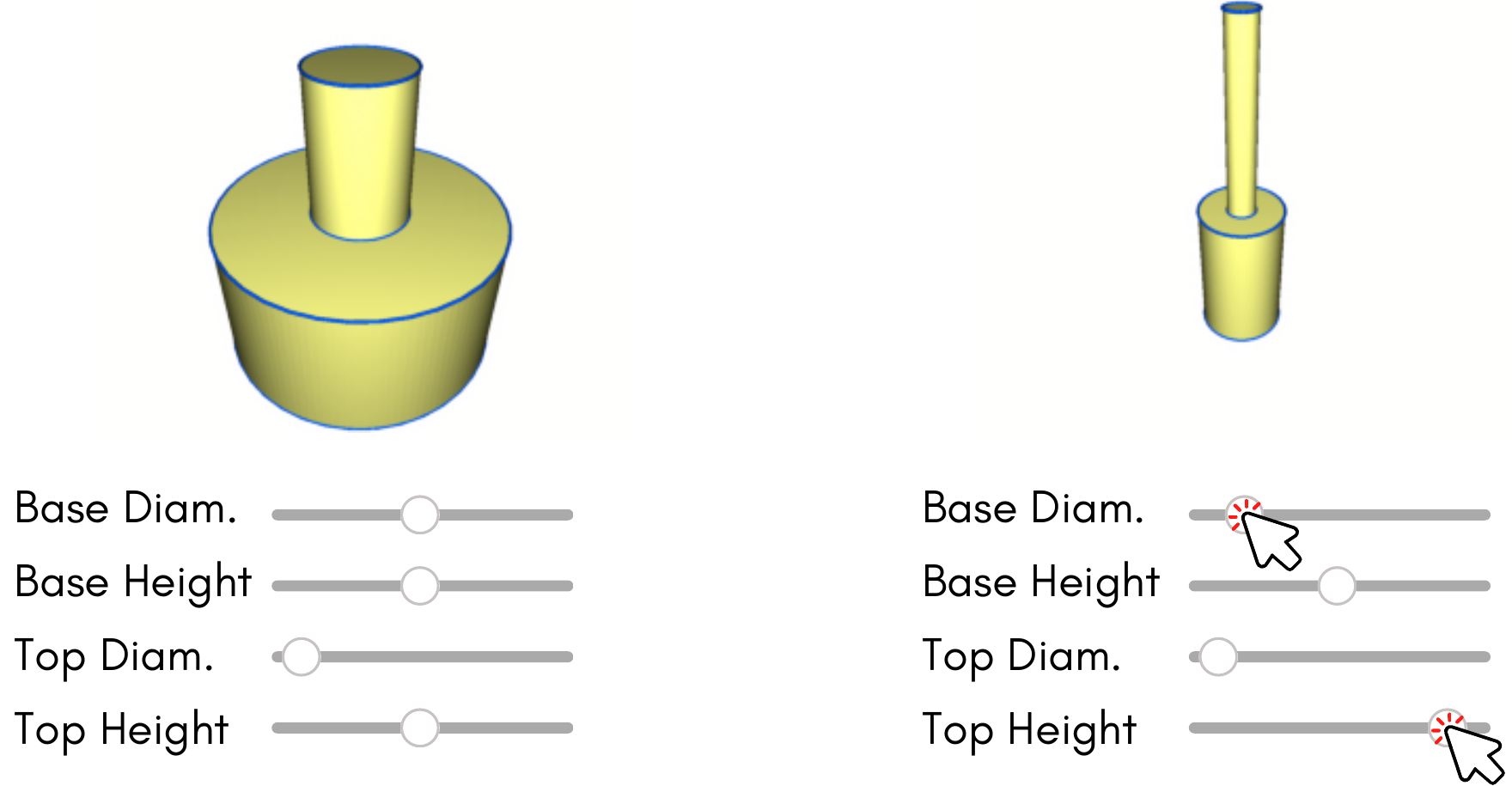}
        \caption{Executed code produces interactive sliders that the designer can use to modify the size of the primitives.}
        \label{fig:editing_exec}
    \end{subfigure}

    \caption{Editing pipeline: given a predicted code from \modelname~from a point cloud (a), a generic prompt can be constructed to refactor the predicted code to enhance editibility (b). The output from GPT-4o-2024-08-06 is shown in (c), and the generated sliders and possible CAD edits are depicted in (d).}
    \label{fig:editing_pipeline}
\end{figure*}

We provide more details on the editing pipeline presented in Section~\textcolor{cvprblue}{5.2} of the main paper. The goal of this pipeline is to integrate automated editability capabilities to \modelname. To this end, we present a simple process using an off-the-shelf LLM, GPT-4o~\cite{openai2024gpt4technicalreport}. Starting from an output CAD Python code from \modelname~as shown in \Cref{fig:editing_code}, we prepare a simple and generic prompt (\Cref{fig:editing_prompt}) for the LLM to generate a refactored version of the code such that when executed the user can change with the dimensions of each primitive. As seen in \Cref{fig:editing_output}, the LLM is able to generate a code with comments that describe the different primitives semanticallly and include appropriate variables for the dimensions of each of the primitive, such as the height and the diameter of each cylinder. The code generated by the LLM, can be directly executed in a Jupyter notebook with the CadQuery and ipywidgets libraries. Figure~\textcolor{cvprblue}{6} of the main paper shows the generated sliders and how can the shape be then edited. This demonstrates that the CAD representation as Python code within a reverse engineering scenario opens the door to new applications when combined with LLMs.

\twocolumn[{}]
{
    \small
    \bibliographystyle{ieeenat_fullname}
    \bibliography{main}
}
\end{document}